\documentclass{article}

\usepackage{microtype}
\usepackage{graphicx}
\usepackage{subfigure}
\usepackage{booktabs} 
\usepackage{paralist}
\usepackage{enumitem}
\usepackage{adjustbox}
\usepackage{sidecap}

\usepackage{import}

\usepackage[nohyperref,accepted]{icml2020}

\usepackage[utf8]{inputenc} 
\usepackage[T1]{fontenc}    
\usepackage{url}

\usepackage{amsmath,amsfonts,bm}

\def\eqref#1{equation~\ref{#1}}

\def\1{\bm{1}}

\def\rvu{{\mathbf{i}}}

\def\rvu{{\mathbf{u}}}

\DeclareMathAlphabet{\mathsfit}{\encodingdefault}{\sfdefault}{m}{sl}
\SetMathAlphabet{\mathsfit}{bold}{\encodingdefault}{\sfdefault}{bx}{n}

\DeclareMathOperator*{\argmin}{arg\,min}

\usepackage{booktabs}       
\usepackage{amsthm}
\usepackage{amsfonts, amssymb}       
\usepackage{nicefrac}       
\usepackage{microtype}      
\usepackage{amsmath}
\usepackage{dsfont}
\usepackage{setspace}
\usepackage{todonotes}
\usepackage[group-separator={,}]{siunitx}

\newcounter{note}

\icmltitlerunning{Learning composable energy surrogates for PDE order reduction}

\begin{document}

\twocolumn[
\icmltitle{Learning Composable Energy Surrogates for PDE Order Reduction}




\begin{icmlauthorlist}
\icmlauthor{Alex Beatson}{cos}
\icmlauthor{Jordan T. Ash}{cos}
\icmlauthor{Geoffrey Roeder}{cos}
\icmlauthor{Tianju Xue}{cee}
\icmlauthor{Ryan P. Adams}{cos}
\end{icmlauthorlist}

\icmlaffiliation{cos}{Department of Computer Science, Princeton University, Princeton, USA}
\icmlaffiliation{cee}{Department of Civil and Environmental Engineering, Princeton University, Princeton, USA}
\icmlcorrespondingauthor{Alex Beatson}{abeatson@cs.princeton.edu}

\icmlkeywords{Machine Learning, Optimization, Bilevel Optimization, Metamaterials, Differential Equations, PDE, Dimension Reduction, Amortization}

\vskip 0.3in
]



\printAffiliationsAndNotice{}  

\begin{abstract}
  Meta-materials are an important emerging class of engineered materials in which complex macroscopic behaviour--whether electromagnetic, thermal, or mechanical--arises from modular substructure.
  Simulation and optimization of these materials are computationally challenging, as rich substructures necessitate high-fidelity finite element meshes to solve the governing PDEs.
  To address this, we leverage \emph{parametric} modular structure to learn component-level surrogates, enabling cheaper high-fidelity simulation.
  We use a neural network to model the stored potential energy in a component given boundary conditions. This yields a structured prediction task: macroscopic behavior is determined by the minimizer of the system's total potential energy, which can be approximated by composing these surrogate models. Composable energy surrogates thus permit simulation in the reduced basis of component boundaries. Costly ground-truth simulation of the full structure is avoided, as
training data are generated by performing finite element analysis with individual components. Using dataset aggregation to choose training boundary conditions allows us to learn energy surrogates which  produce accurate macroscopic behavior when composed, accelerating simulation of parametric meta-materials.



\end{abstract}

\section{Introduction}
Many physical, biological, and mathematical systems are most successfully modeled by partial differential equations (PDEs). 
Analytic solutions are rarely available for PDEs of practical importance; thus, computational methods to approximate PDE solutions are critical for tackling many problems in science and engineering.
Finite element analysis (FEA) is one of the most widely used techniques for solving PDEs on spatial domains; the continuous problem is discretized and replaced by basis functions on a mesh.

The effectiveness of FEA and related methods is largely governed by the granularity of the discrete approximation, i.e., the fineness of the mesh.
Complicated domains can require fine meshes that result in prohibitively expensive computations to solve the PDE.
This problem is compounded when the task is one of parameter identification or design optimization. In these situations the PDE must be repeatedly solved in the inner loop of a bi-level optimization problem.

An important domain where this challenge is particularly relevant is in the modeling of mechanical meta-materials.
Meta-materials are solids in which heterogeneous microstructure leads to rich spaces of macroscopic behavior.
Meta-materials offer the possibility of achieving electromagnetic and/or mechanical properties that are otherwise impossible with homogenous materials and traditional design approaches \citep{poddubny2013hyperbolic,cai2010optical,bertoldi2017flexible}.
In this paper we focus on the class of \emph{cellular} mechanical meta-materials proposed by \citet{overvelde2014relating}, which promise new high-performance materials for soft robotics and other domains (see Sec 3).
The simulation of these meta-materials is particularly challenging due to the complex cellular structure and the need to accurately capture small-scale nonlinear elastic behavior.
Traditional finite element methods have limited ability to scale to these problems, and automated design of these promising materials demands accurate and efficient approximate solutions to the associated PDE.

We develop a framework which takes advantage of spatially local structure in large-scale optimization problems---here the minimization of energy as a function of meta-material displacements. 
Given that only a small subset of material displacements are of interest, we ``collapse out'' the remainder using a learned surrogate.
Given a component with substructure defined by local parameters, the surrogate produces an accurate proxy energy in terms of the boundary alone.
A single surrogate can be trained, then replicated and composed to predict the energy of a larger solid---which is the sum of energies over sub-components.
This allows solving the PDE in the reduced basis of component boundaries by minimizing the sum of surrogate energies.

Other methods exist for amortizing the solution of PDEs. 
Some of the most common approaches use neural networks to map from PDE parameters to solutions \citep{zhu2019physics,nie2020stress} or construct reduced bases via solving eigenvalue problems or interpolating between snapshots \citep{berkooz1993proper, chatterjee2000introduction}. 
These approaches typically require solving full systems to produce training data. 
Our framework uses the modular decomposition of energy to train surrogate models on data generated by querying the finite element "expert" on the energy in small components, avoiding performing FEA on large systems which are expensive to solve.

The remaining sections of this paper are as follows.
Section~\ref{sec:collapsed} proposes a framework for collapsed-basis optimization when the objective decomposes as a sum of terms which each depend only on local variables. 
Section~\ref{sec:metamaterials} provides a brief introduction to the cellular solids of interest and the PDEs governing their behavior.
In Section~\ref{sec:surrogates}, collapsed-basis optimization is applied to the cellular meta-material domain, using the architecture described in Section~\ref{sec:model}.
The specifics of the training procedure and our use of the imitation learning technique \textsc{DAgger} are explained in Section~\ref{sec:training}. 
Section 7 describes the software and hardware used.
In Section~\ref{sec:results}, empirical evaluation demonstrates that composable energy surrogate models are able to solve cellular solid PDEs accurately and efficiently.
Limitations and future work are discussed in Section~\ref{sec:discussion}.

\section{Learning to optimize in collapsed bases}
\label{sec:collapsed}
Solving PDEs like those that govern meta-material behavior can be framed as an optimization problem of finding a solution~$u$ which minimizes an energy~$E(u)$ subject to constraints.
For mechanical meta-materials, $E(u)$ is the stored elastic potential energy in the material.
We propose a method for amortizing high-dimensional optimization problems where the objective has special conditional independence structure, such as that found in solving these PDEs. Consider the general problem of solving 
\begin{equation}\label{eq:minimization}
u^* = \argmin E(u)\,.
\end{equation}
Here,~$u$ may be a vector in~$\mathbb{R}^d$ or may belong to a richer space of functions.
Often we are only interested in either a small subset of the vector~$u^*$ or the values the function~$u^*$ takes on a small subdomain.
To reflect this structure, we take the solution space to be the Cartesian product of a space of primary interest and a ``nuisance'' space.
We denote the solutions as concatenations~${u=[x,y]}$ where~$y$ is the object of interest, and~$x$ is the object whose value is not of interest to an application.~$x$ is roughly equivalent to auxiliary variables that often appear in probabilistic models, but that are marginalized away or discarded from the simulation.
We can use this decomposition to frame Eq.~\ref{eq:minimization} as a bi-level optimization problem:
\begin{equation}\label{eq:bilevel}
y^* = \argmin_y \min_x E(x, y)\,.
\end{equation}
Consider the \emph{collapsed objective},~${\tilde{E}(y) = \min_x E(x, y)}$.
If it is possible to query~$\tilde{E}(y)$ and its derivatives without ever representing~$x$, we may perform \emph{collapsed optimization} in the reduced basis of~$y$, avoiding either performing optimization in the larger basis of~$u$ (Eq. \ref{eq:minimization}), or performing bi-level optimization (Eq. \ref{eq:bilevel}) with an inner loop.

However,~$\tilde{E}$ is not usually available in closed form.
We might consider approximating~$\tilde{E}(y)$ via supervised learning.
In the general case, this would require solving~${\tilde{E} = \min_x E(x, y)}$ for each example~$y$ we wish to include in our training set.
This is the procedure used by many surrogate-based optimization techniques \citep{queipo2005surrogate,forrester2009recent,shahriari2015taking}.
The high cost of gathering each training example makes this prohibitive when~$x$ is high dimensional (such that the minimization is difficult) or when~$y$ is high dimensional (such that fitting a surrogate requires many examples).

In some cases, compositional structure in~$E$ may assist us with efficiently approximating~$\tilde{E}$.
Many objectives may be represented as a sum:
\begin{equation}\label{eq:sum}
    E(x, y) = \sum_i E_i(x_i, y)\,.
\end{equation}
Given this decomposition,~$x_i$ and~$x_j$ are conditionally independent given~$y$; i.e., if we constrain~$x_i$ and~$y$ to take some values and perform minimization, the  resulting $x_j$ or $E_j(x_j, y)$ do not vary with the value chosen for $x_i$. 
This follows from the partial derivative structure~${\frac{\partial{E_i}}{\partial x_j} = 0}$ for~${i\neq j}$.

We propose to \emph{learn} a collapsed objective $\tilde{E}$, which exploits conditional independence structure by representing~${\tilde{E}(y) = \sum_i \tilde{E}_i(y)}$. 
This representation as a sum allows us to use~${\min_{x_i} E_i(x_i, y)}$ as targets for supervision, which may be found more cheaply than performing a full minimization.
The learned approximations to~$\tilde{E}_i$ may be composed to form an energy function with larger domain.

The language we use to describe this decomposition is intentionally chosen to reflect the conceptual similarity of our framework to \emph{collapsed variational inference} \citep{teh2007collapsed} and \emph{collapsed Gibbs sampling} \citep{geman1984stochastic, liu1994collapsed}.
In these procedures, conditional independence is exploited to allow optimization or sampling to proceed in a collapsed space where nuisance random variables are marginalized out of the relevant densities.
We exploit similar structure to these techniques, albeit in a deterministic setting.
Other approaches to accelerating Eq.~\ref{eq:bilevel} which do not exploit (\ref{eq:sum}) or directly model~$\tilde{E}(y)$ include amortizing the inner optimization by predicting~${x^*(y) = \argmin_x E(x, y)}$ \citep{kingma2013auto,brock2017smash}, or truncation of the inner loop, either deterministic \citep{wu2018understanding,shaban2018truncated} or randomized to reduce bias \citep{tallec2017unbiasing,beatson2019efficient}. 

The specific optimization procedure we accelerate is the numerical simulation of mechanical materials, where the objective corresponds to a physically meaningful energy, and the conditional independence structure arises from a spatial decomposition of the domain and the spatial locality of the energy density.
We believe this spatial decomposition of the domain and energy could be generalized to learn collapsed energies for solving many other PDEs in reduced bases.
This collapsed-basis approach may also be applicable to other bi-level optimization problems where the objective decomposes as a sum of local terms.

\section{Mechanical meta-materials}
\label{sec:metamaterials}
Meta-materials are engineered materials with microstructure which results in macroscopic behavior not found in nature. 
They are often discussed in the context of materials achieving specific electromagnetic phenomena, such as negative refraction index solids and ``invisibility cloaks'' which electromagnetically conceal an object through engineered distortion of electromagnetic fields \citep{poddubny2013hyperbolic,cai2010optical}. 
However, they also hold great promise in other domains: \emph{mechanical} meta-materials use substructure to achieve unusual macroscopic behavior such as negative Poisson's ratio and nonlinear elastic responses; pores and lattices undergo reversible collapse under large deformation, enabling the engineering of complex physical affordances in soft robotics \citep{bertoldi2017flexible}. 

 		\begin{figure}
 		\begin{tabular}{c|c|c}
    		 {{\includegraphics[height=2.2cm, trim={3.5cm 3.5cm 3.5cm 3.5cm}, clip]{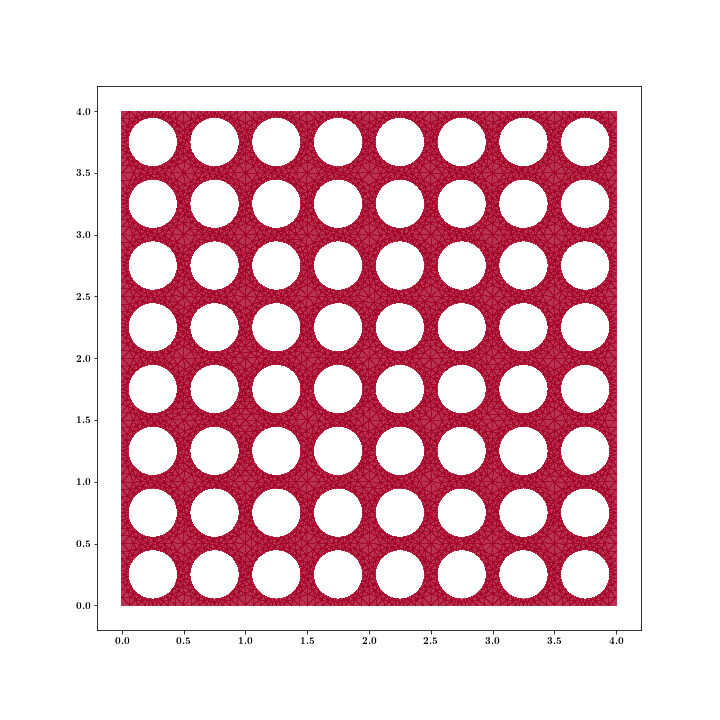} }}&
    		{{\includegraphics[height=2.2cm, trim={3.5cm 3.5cm 3.5cm 3.5cm}, clip]{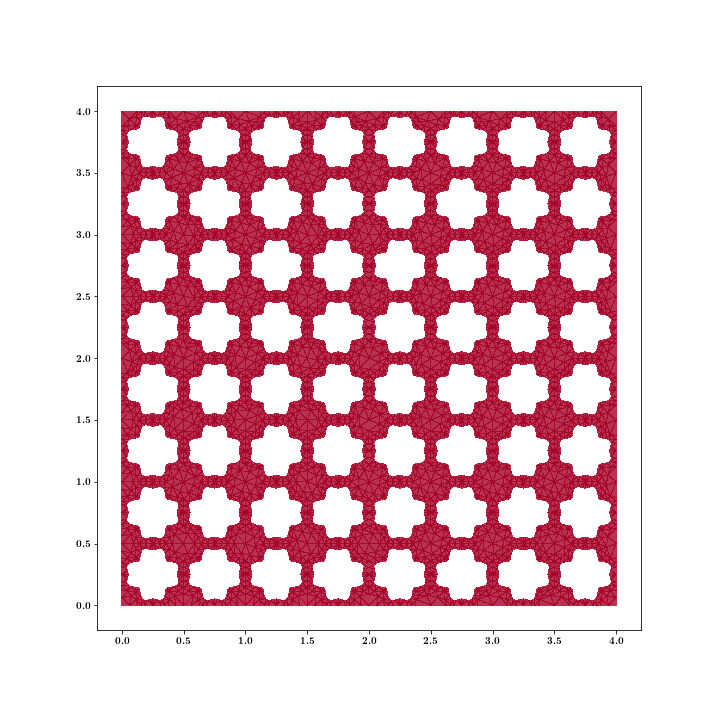} }}&%
    		{{\includegraphics[height=2.2cm, trim={3.5cm 3.5cm 3.5cm 3.5cm}, clip]{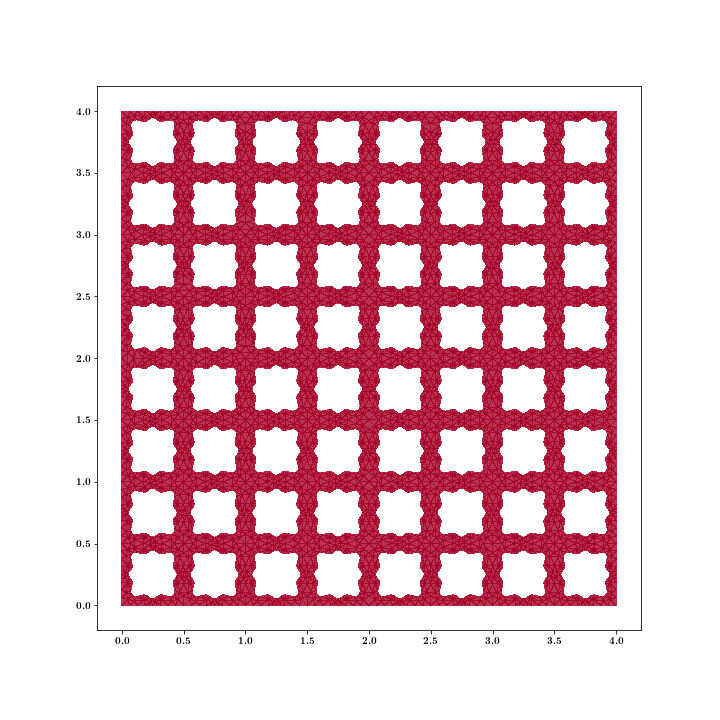} }}\\\hline  
    	
    	\rule{0pt}{13ex}{{\includegraphics[height=1.9cm, trim={3.5cm 4.5cm 3.5cm 4.5cm}, clip]{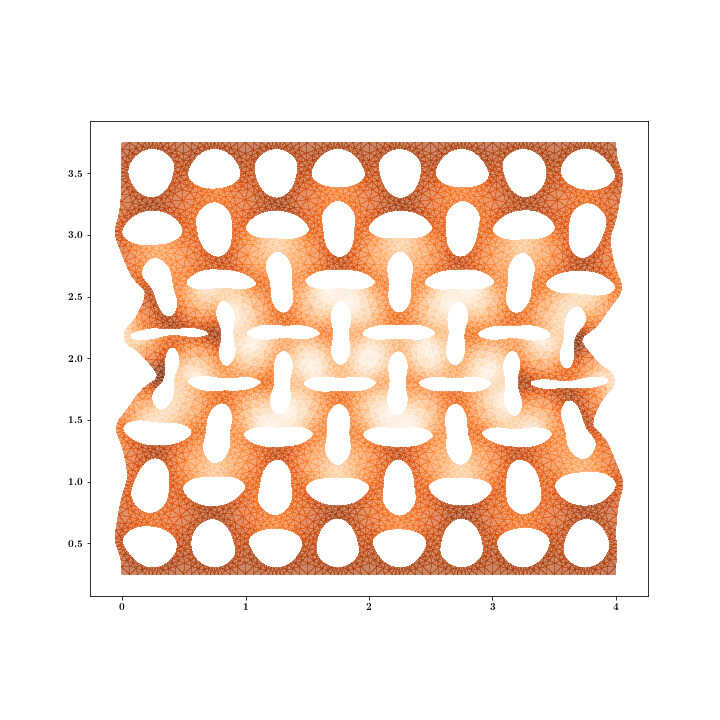} }}&%
    		{{\includegraphics[height=1.9cm, trim={3.5cm 4.5cm 3.5cm 4.5cm}, clip]{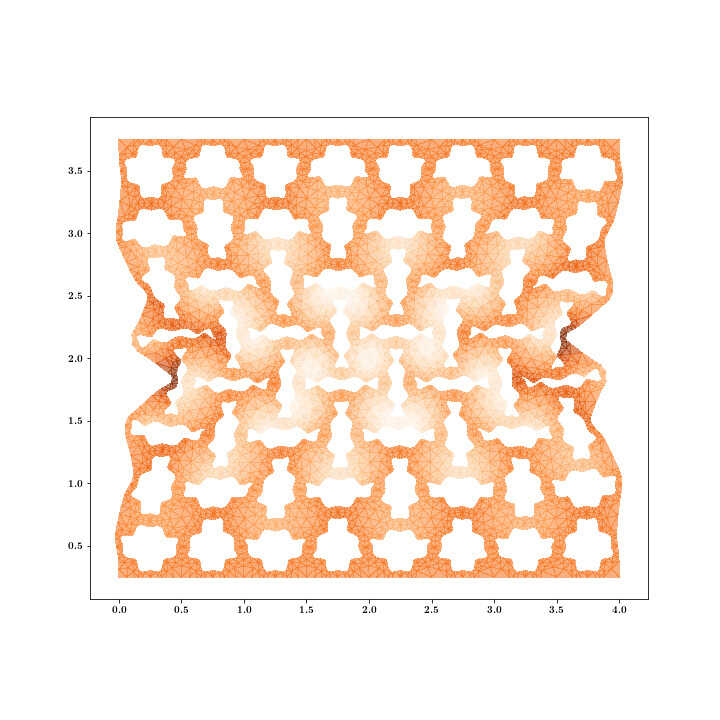} }}&%
    		{{\includegraphics[height=1.8cm, trim={3.5cm 6cm 3.5cm 6cm}, clip]{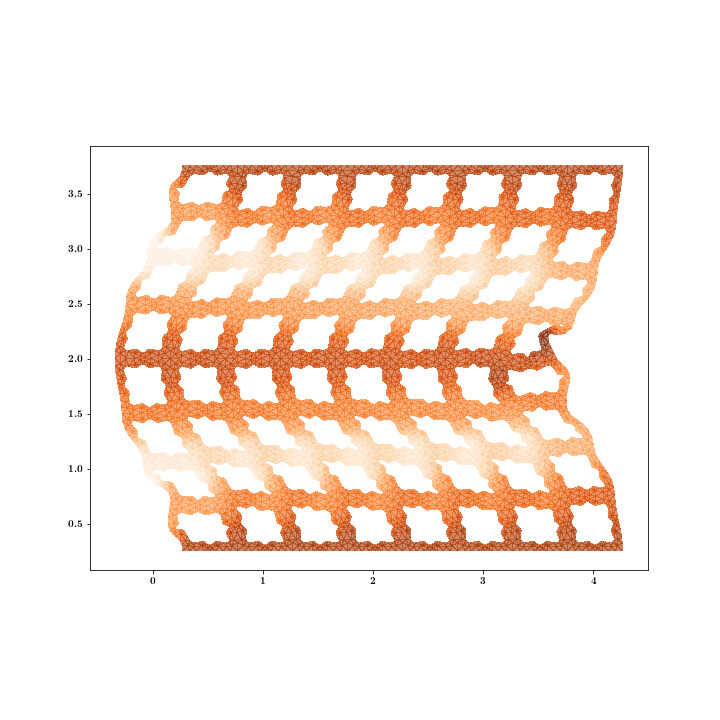} }}%
    	\end{tabular}
    		\vspace{-0.25cm}%
    		\caption{Cellular meta-materials. Top: materials at rest. Bottom: materials under axial compression, exhibiting periodic instability that varies with pore shape. The left two structures exhibit a negative Poisson's ratio, which does not occur in nature.}%
    		\label{Fig:mms}%
		\vspace{-0.5cm}
		\end{figure} 

Meta-materials hold promise for modern engineering design, but are challenging to simulate as the microstructure necessitates a very fine finite element mesh, and as the nonlinear response makes them difficult to approximate with a macroscopic material model. 
Most work on meta-materials has thus relied on engineers and scientists to hand-design materials, rather than numerically optimizing meta-material substructure to maximize some objective \citep{ion2016metamaterial}.

We focus on building surrogate models for the two-dimensional cellular solids investigated in \citet{overvelde2014relating}.
These meta-materials consist of square ``cells'' of elastomer, each of which has a pore in its center. 
The pore shapes are defined by parameters~$\alpha$ and~$\beta$ which characterize the pore shape in polar coordinates:
\begin{equation}
r(\theta) = r_0[1 + \alpha \cos(4\theta) + \beta \cos(8\theta)]
\end{equation}
$r_0$ is chosen such that the pore covers half the cell's volume:~${r_0 = \frac{L_0}{\sqrt{\pi(2 + \alpha^2 + \beta^2}}}$. Constraints are placed on~$\alpha$ and~$\beta$ to enforce a minimum material thicknesses and ensure that $\min_\theta r(\theta) > 0$ as in \citet{overvelde2014relating}.

These pore shapes give rise to complicated nonlinear elastic behavior, including negative Poisson's ratio and double energy wells (i.e., stored elastic energy which does not increase monotonically with strain).
Realizations of this class of materials are shown under axial strain in Figure \ref{Fig:mms}. 

The continuum mechanics behavior of these elastomer meta-materials can be captured by a neo-Hookean energy model \citep{ogden1997non}. 
Let~${X \in \mathbb{R}^d}$, where~${d\leq 3}$ in physical problems, be a point in the resting undeformed material reference configuration, and~$u(X)$ be the displacement of this point from reference configuration. The stored energy in a neo-Hookean solid is
\begin{equation}
E = \int_{\Omega} W(u) dX\,,
\end{equation}
where $W(u)$ is a scalar energy density over $\Omega$. It is defined for material bulk and shear moduli~$\mu$ and~$\kappa$ as:
\begin{equation}\label{eq:neohookean}
   W = \frac{\mu}{2} \Big( (\det F)^{-2/d} \text{tr}(FF^T) - d\Big) + \frac{\kappa}{2} (\det F - 1)^2
\end{equation}
where $F$ is the deformation gradient:
\begin{equation}
F(X) = \frac{\partial u(X)}{\partial X} + I\,.
\end{equation}
The pore shapes influence the structure of these equations by changing the material domain $\Omega$.
The behaviour of these solids can be simulated by solving:
\begin{equation}\label{eq:cm}
    \text{Div } S = 0 \quad X \in \Omega
\end{equation}
\vspace{-0.1cm}
\begin{equation}\label{eq:bc}
    G(u) = 0 \quad X \in \partial \Omega
\end{equation}
where~${S = \frac{\partial W}{\partial F}}$ is known as the first Piola-Kirchoff stress, and where Eq.~\ref{eq:bc} defines a boundary condition.
For example,~${G(u) = u - u_b}$ corresponds to a Dirichlet boundary condition; in our case, an externally imposed displacement. ${G(u) = \frac{\partial W}{\partial u} - f_b}$ corresponds to an external force exerting a pressure on the boundary.

To simulate these meta-materials, the PDE in Eq.~\ref{eq:cm} is typically solved via finite element analysis.
Solving the PDEs arising from large mechanical meta-material structures is computationally challenging due to fine mesh needed to capture pore geometry and due to the highly nonlinear response induced by buckling under large displacements.

Solving the PDE in Eq.~\ref{eq:cm} corresponds to finding the~$u$ which minimizes the stored energy in the material subject to boundary conditions.
That is, Eqs.~\ref{eq:cm} and~\ref{eq:bc} may be equivalently be expressed in an energy minimization form:
\begin{equation}\label{eq:cm-energy}
    		u = \argmin \int_{X \in \Omega}  W(u) dX
\end{equation}
\begin{equation}\label{eq:bc-energy}
    		\text{subject to } G(u) = 0 \in \partial\Omega
\end{equation}
In the next section, we use this energy-based perspective to facilitate learning energy surrogates which allow solving the PDE in a reduced basis of meta-material cell boundaries.

\begin{SCfigure}[1.1]
\includegraphics[height=0.5\linewidth, trim={4.0cm 5cm 3.5cm 4.5cm}, clip]{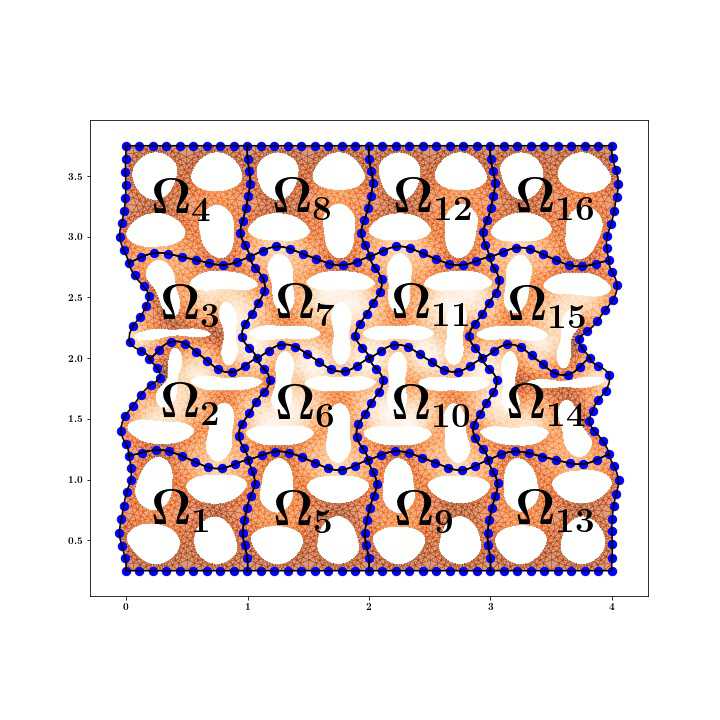}

	\caption{Partitioning a meta-material domain $\Omega$, shown under axial compression, into components $\Omega_1$ to $\Omega_{16}$.
	Black lines show the skeleton $\mathcal{B}$.
	Blue points locate control points of the splines used to represent the reduced-basis solution $\tilde{u}$.\vspace{-.5cm}}
	\label{Fig:decomp}
\vspace{-.5cm}
\end{SCfigure}
\section{Composable energy surrogates}
\label{sec:surrogates}
We apply the idea of learning collapsed objectives to the problem of simulating two-dimensional cellular mechanical meta-material behavior.
The material response is determined by the displacement field $u$ which minimizes the energy~${\int_{\Omega} W dX}$, subject to boundary conditions. 
We divide $\Omega$ into regular square subregions $\Omega_i$, which we choose to be cells with $2\times2$ arrays of pores, and denote the intersection of the subregion boundaries with~${\mathcal{B} = \partial\Omega_1 \cup \partial\Omega_2 \cup} \dots$
We let $u_i$ be the restriction of $u$ to $\Omega_i$.
We take the quantity of interest to be $u_{\mathcal{B}}$, the restriction of $u$ to $\mathcal{B}$, and the nuisance variables to be the restriction of $u$ to~${\Omega \textbackslash \mathcal{B}}$.
The partitioning of $\Omega$ is shown in Figure \ref{Fig:decomp}.

The total energy in the material decomposes as a sum over the energy within each region:
\begin{align*}
    E(u) = &\int_{X \in \Omega} W(u) dX \\
    = \sum_i &\int_{X \in \Omega_i} W(u_i) dX := \sum_i E(u_i) 
\end{align*}

Let $\tilde{u}_i$ be the restriction of $u$ to $\partial\Omega_i$, noting~${\partial\Omega_i = \mathcal{B} \cap \Omega_i}$. 
We introduce the collapsed energy of a component:
\begin{align*}
    \tilde{E}_i (\tilde{u}_i) :&= \min_{u_i} E (u_i)\\
    \text{subject to } u_i(X) &= \tilde{u}_i(X)\quad X \in \partial{\Omega_i}\,.
\end{align*}
This quantity is the lowest energy achievable by displacements of the \emph{interior} of the cell~$\Omega_i$, given the boundary conditions specified by~$\tilde{u}_i$ on~$\partial{\Omega_i}$.
$\tilde{E}_i(\tilde{u}_i)$ depends on the shape of the region $\Omega_i$, i.e., on the geometry of the pores. 
Rather than each possible pore shape having a unique collapsed energy function, we introduce the pore shape parameter~${\xi = (\alpha, \beta)}$ as an argument, replacing~$\tilde{E}_i(\tilde{u_i})$ with~$\tilde{E}(\tilde{u}_i, \xi_i)$.
The macroscopic behavior of the material is fully determined by this \emph{single} collapsed energy function~${\tilde{E}(\tilde{u}_i, \xi_i)}$. 
Given the true collapsed energy functions, we could accurately simulate material behavior in the reduced basis of the boundaries between each component~$\Omega_i$.\footnote{So long as forces and constraints are only applied on $\mathcal{B}$.}

We learn to approximate this collapsed energy function from data. This function may be duplicated and composed to simulate the material in the reduced basis~$\mathcal{B}$, an approach we term \emph{composable energy surrogates} (CESs). 
A single CES is trained to approximate the function $\tilde{E}$ by fitting to supervised data~${(\tilde{u}_i, \xi_i, \tilde{E}(\tilde{u}_i, \xi_i))}$, where $\xi_i$ and $\tilde{u}_i$ may be drawn from any distribution corresponding to anticipated pore shapes and displacements, and the targets $\tilde{E}(\tilde{u}_i, \xi_i)$ are generated by solving the PDE in a small region $\Omega_i$ with geometry defined by $\xi_i$ and with $\tilde{u}_i$ imposed as a boundary condition. 
The CES may be duplicated and composed to approximate the total energy of larger cellular meta-materials.

Our ultimate goal is to efficiently solve for a reduced-basis displacement~$u_{\mathcal{B}}$ on~$\mathcal{B}$.
Reduced-basis solving via CES may be cast as a highly-structured imitation learning problem. 
Consider using a gradient-based method to minimize the composed surrogate energy:
\begin{equation}
    \hat{E}(u_\mathcal{B}) = \sum_i \hat{E}(\tilde{u}_i, \xi_i)
\end{equation}
where~$\hat{E}(\tilde{u}_i, \xi_i)$ is the model's prediction of $\tilde{E}(\tilde{u}_i, \xi_i)$, the collapsed energy of a single component.
A sufficient condition for finding the correct minimum is for the "action" taken by the surrogate---the derivative of the energy approximation~${\nabla_{u_{\mathcal{B}}} \hat{E}}$---to match the "action" taken by an expert---the \emph{total} derivative,~${\nabla_{u_{\mathcal{B}}} \min_{u\notin \mathcal{B}} E(u)}$---along the optimization trajectory. 
If so, the surrogate will follow the trajectory of a valid, if non-standard, bilevel gradient-based procedure for minimizing the energy, corresponding to (\ref{eq:bilevel}). Given an imperfect surrogate, the error in the final solution could trivially be bounded in terms of the error in approximating~${\nabla_{u_{\mathcal{B}}} \min_{u\notin \mathcal{B}} E(u)}$ by $\nabla_{u_{\mathcal{B}}} \hat{E}$ along the trajectory, and the number of gradient steps taken. 
This observation informs our model, training, and data collection procedures, described in the following sections.

\setlength{\abovedisplayskip}{6pt}
\setlength{\belowdisplayskip}{5pt}
\section{Model architecture}
\label{sec:model}
Our CESs take the form of a neural architecture, designed to respect known properties of the true potential energy and to maximize usefulness as surrogate energy to be minimized via a gradient-based procedure. 
The effects of these design choices are quantified via an ablation study in the appendix.

\textbf{Reduced-basis parameterization}.
We require a vector representation for the function $\tilde{u}$.
We use one cubic spline for each horizontal and vertical displacement function along each face of the square, with evenly spaced control points and ``not-a-knot'' boundary conditions. 
Our vector representation is~${\rvu \in \mathbb{R}^{2n}}$, formed from the horizontal and the vertical displacement values at each of the $n$ control points.
Splines on adjacent faces share a control point at the corner. Using $N$ control points to parameterize the function along each face requires~${n = 4*(N-1)}$ control points to parameterize a $1d$ function around a single cell. 
For all experiments we use~${N = 10}$ control points along each edge, resulting in a vector $\rvu$ with 72 entries.

\textbf{Model structure and loss}. Our surrogate energy model is:
\begin{equation*}
    \hat{E}(\rvu, \xi) = \underbrace{||\mathcal{R}(\rvu)||_2^2}_\text{Linear elastic component} \underbrace{\exp \{f_\phi \big(\mathcal{R}(\rvu), \xi\big)\}}_\text{Stiffness}\,,
\end{equation*}
where $f_\phi$ is a neural network with parameters $\phi$ and $\mathcal{R}$ removes rigid-body rotation and translation.
Our loss function is~${\mathcal{L} =  \mathcal{L}^0 + \mathcal{L}^1 + \mathcal{L}^2\,,}$ 
which is the sum of losses on the $0$th, $1$st and $2$nd energy derivatives:
\begin{equation*}
    \mathcal{L}^0 = \underbrace{\left\lVert f_\phi \big(\mathcal{R}(\rvu), \xi\big) - \log \frac{\tilde{E}(\tilde{u})}{||\mathcal{R}(\rvu)||_2^2} \right\rVert_2^2}_\text{Log-stiffness loss} 
\end{equation*}
\begin{equation*}
    \mathcal{L}^1 = \underbrace{1 - \frac{\langle \nabla_{\rvu} \hat{E}, \nabla_{\rvu} \tilde{E}\rangle}{||\nabla_{\rvu} \hat{E}|| ||\nabla_{\rvu} \tilde{E}||}}_\text{Cosine distance between gradients}
\end{equation*}
\begin{equation*}
\mathcal{L}^2 = \underbrace{1 - \frac{\langle \nabla^2_{\rvu} \hat{E} v,  \nabla^2_{\rvu} \tilde{E} v \rangle}{||\nabla^2_{\rvu} \hat{E} v|| ||\nabla^2_{\rvu} \tilde{E} v||}}_{\substack{\text{Cosine distance between}\\\text{Hessian-vector products}}} \quad \underbrace{v \sim \mathcal{N}(0, I^{2n})}_\text{Projection vector for Hessian}\,.
\end{equation*}
In the above, $\nabla_{\rvu}$ and $\nabla^2_{\rvu}$ are the gradient and Hessian of the surrogate energy $\hat{E}$ or the ground-truth energy $\tilde{E}$ with respect to $\rvu$, and $v$ is sampled independently for each training example in a batch. 

\textbf{Invariance to rigid body transforms}. 
The true stored elastic energy is invariant to rigid body transformation of a solid. 
This invariance may be hard for a neural network to learn exactly from data. We define a module $\mathcal{R}$ which applies \emph{Procrustes analysis}, a procedure that finds and applies the rigid body transformation which minimizes the Euclidean distance to a reference, for which we use the rest configuration. This is differentiable and closed-form.

\textbf{Encoding a linear elastic bias}.
The energy is approximated well by a linear elastic model when at rest:~${\tilde{E}^i(\tilde{u}_i) \approx \mathcal{R}(\rvu_i)^T A^i \mathcal{R}(\rvu_i)}$ for a stiffness matrix~$A^i$ depending on~$\xi_i$. 
We scale our net's outputs by~$||\mathcal{R}(\rvu_i)||_2^2$ so that it needs only capture a ``scalar stiffness''~$\nicefrac{E}{||\mathcal{R}(\rvu_i)||_2^2}$ accounting for the geometry of~$A^i$ given~$\xi_i$ and for deviation from the linear elastic model.

\textbf{Parameterizing the log-stiffness}.
The energy of a component $\tilde{E}^i(u_{0, i})$ is nonnegative, and the ratio of energy to a linear elastic approximation varies over many orders of magnitude. 
We thus parameterize the log of the scalar stiffness with our neural network $f_\phi$ rather than the stiffness.

\textbf{Log-stiffness loss}. We wish to find neural network parameters $\phi$ which lead to accurate energy predictions for many different orders of magnitude of energy and displacement. 
Minimizing the~$\ell^2$ loss between predicted and true energies penalizes errors in predicting large energies more than proportional errors predicting small energies. 
Instead, we take the~$\ell^2$ loss between the predicted log-stiffness~$f_\phi (\mathcal{R}(\rvu), \xi)$ and the effective ground-truth log-stiffness,~$\log\nicefrac{ \tilde{E}(\tilde{u})}{||\mathcal{R}(\rvu)||_2^2}$.

\textbf{Sobolev training with gradients}. 
When derivatives of a target function are available, training a model to match these derivatives (``Sobolev training'') can aid generalization \citep{czarnecki2017sobolev}.
Accuracy of CES' gradients is crucial to an accurate solution trajectory.
We obtain ground-truth gradients cheaply via the adjoint method \citep{lions1971optimal}. Given a solution $u_i$ to the PDE in $\Omega_i$ with boundary conditions $\tilde{u}_i$, the gradient~${\nabla_{\tilde{u}_i} \tilde{E}_i(\tilde{u}_i)}$ requires one linear system solve, with the same cost as one Newton step while solving the PDE \citep{mitusch2019dolfin}. 
The spline is a linear map $\mathcal{M}$ from $\rvu_i$ to $\tilde{u}_i$ in the finite element basis. 
We can thus efficiently compute~${\nabla_{\rvu_i} \tilde{E}_i(\tilde{u}_i) = \mathcal{M}^T \nabla_{\tilde{u}_i} \tilde{E}_i(\tilde{u}_i)}$. The gradient of our model,~${\nabla_{\rvu_i} \hat{E}_\phi(\rvu_i, \xi_i)}$, may be computed with one backward pass.

\textbf{Sobolev training with Hessian-vector products}. 
Given a solution and gradient, computing $\nabla^2_{\rvu}\tilde{E}$ requires $2n$ linear system solves---one for each entry in $\rvu$. As $2n=72$ is much smaller than the number of Newton steps we need to solve the PDE for moderate displacements, we expect the increased fidelity from a 2nd-order approximation of the energy to be worth this added computation.
Computing the full Hessian of the surrogate energy,~${\nabla^2_{\rvu_i}\hat{E}_\phi(\rvu_i, \xi_i)}$, would require $2n$ backward passes. Instead we train on Hessian-vector products, which require only a single backward pass additional to that required for the gradient. 

\textbf{Cosine distance loss for Sobolev training}. 
Ground-truth gradient and Hessian values vary over many orders of magnitude, roughly corresponding to lower and higher energy displacements. 
We wish our model to be robust to outliers and accurate across a range of different conditions. 
Rather than placing an~$\ell^2$ loss on the gradient and Hessian-vector products as in \citet{czarnecki2017sobolev}, we minimize the cosine distance between ground truth and approximate gradients and Hessians, which is naturally bounded in $[0, 1]$.
\vspace{-0.5cm}

\begin{figure*}[h!]
 \begin{tabular}{cc}
	 {\resizebox{0.45\textwidth}{!}{
 \begin{adjustbox}{clip, trim=.4cm .3cm .7cm .3cm}
	 \includegraphics{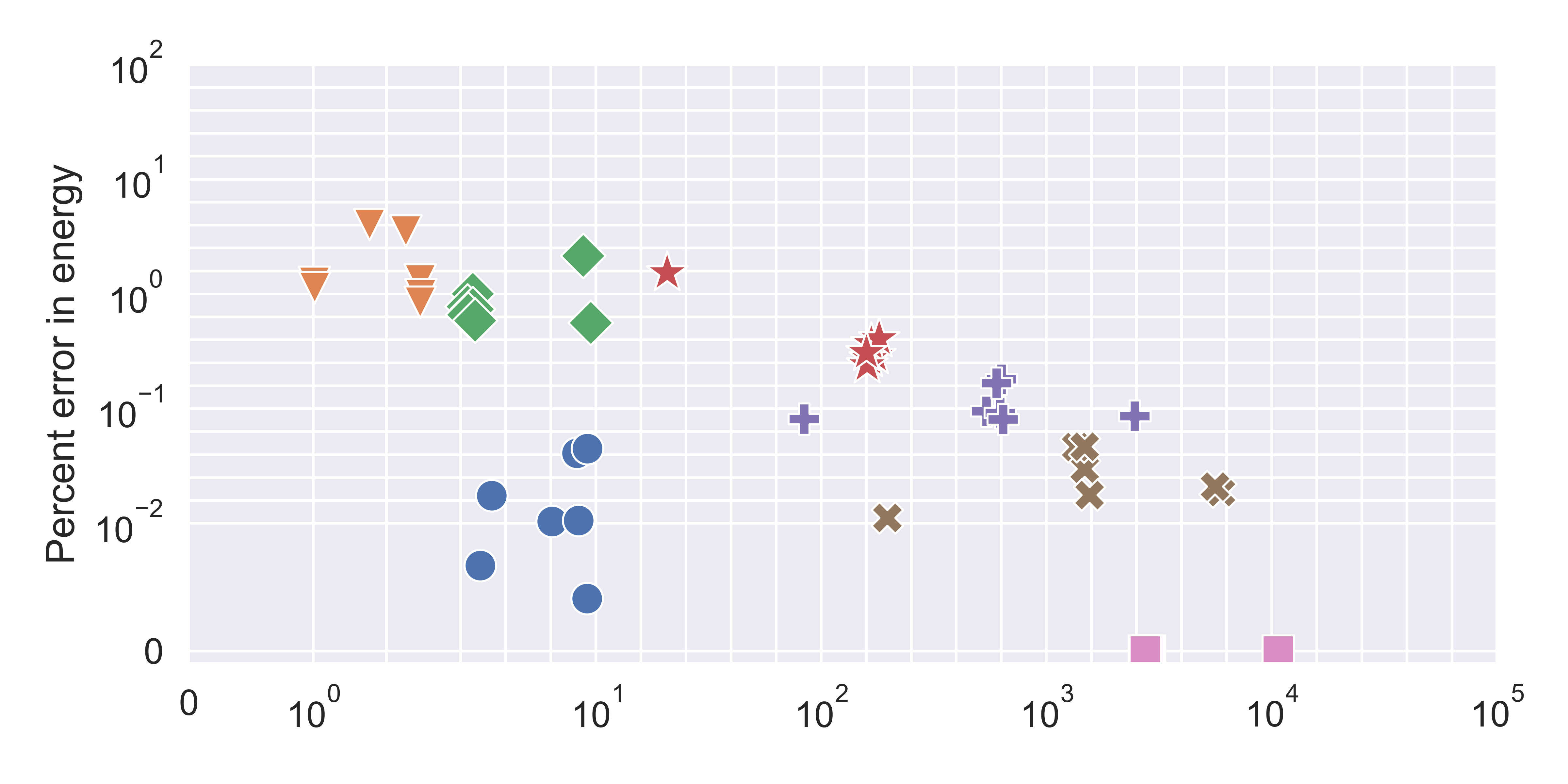}
	 \end{adjustbox}
	 }}&
	{\resizebox{0.45\textwidth}{!}{
 \begin{adjustbox}{clip, trim=.4cm .3cm .7cm .3cm}
	\includegraphics{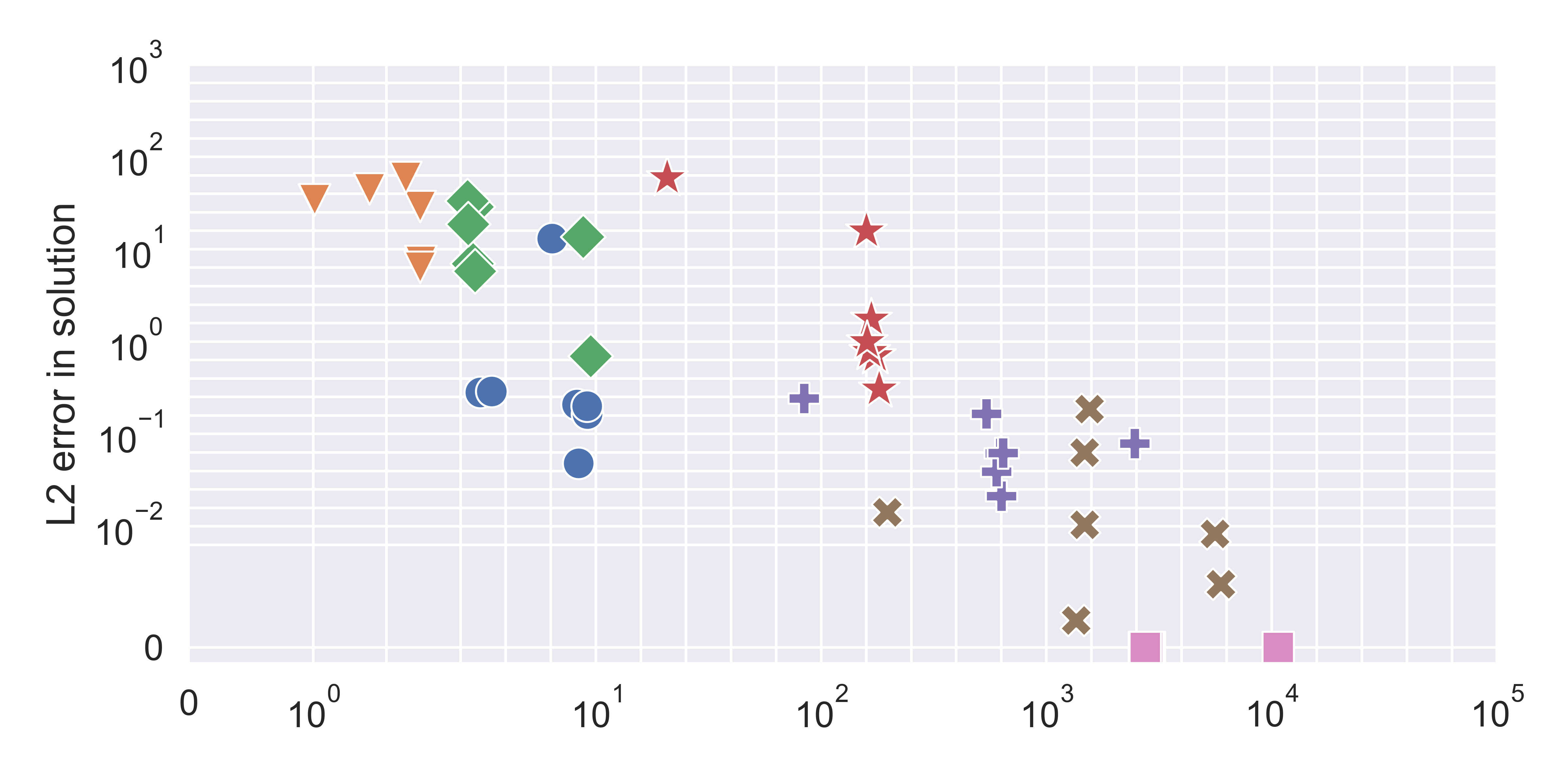}
	 \end{adjustbox}
	}}\\
{\resizebox{0.45\textwidth}{!}{
 \begin{adjustbox}{clip, trim=.3cm 0cm .7cm .3cm}
\includegraphics{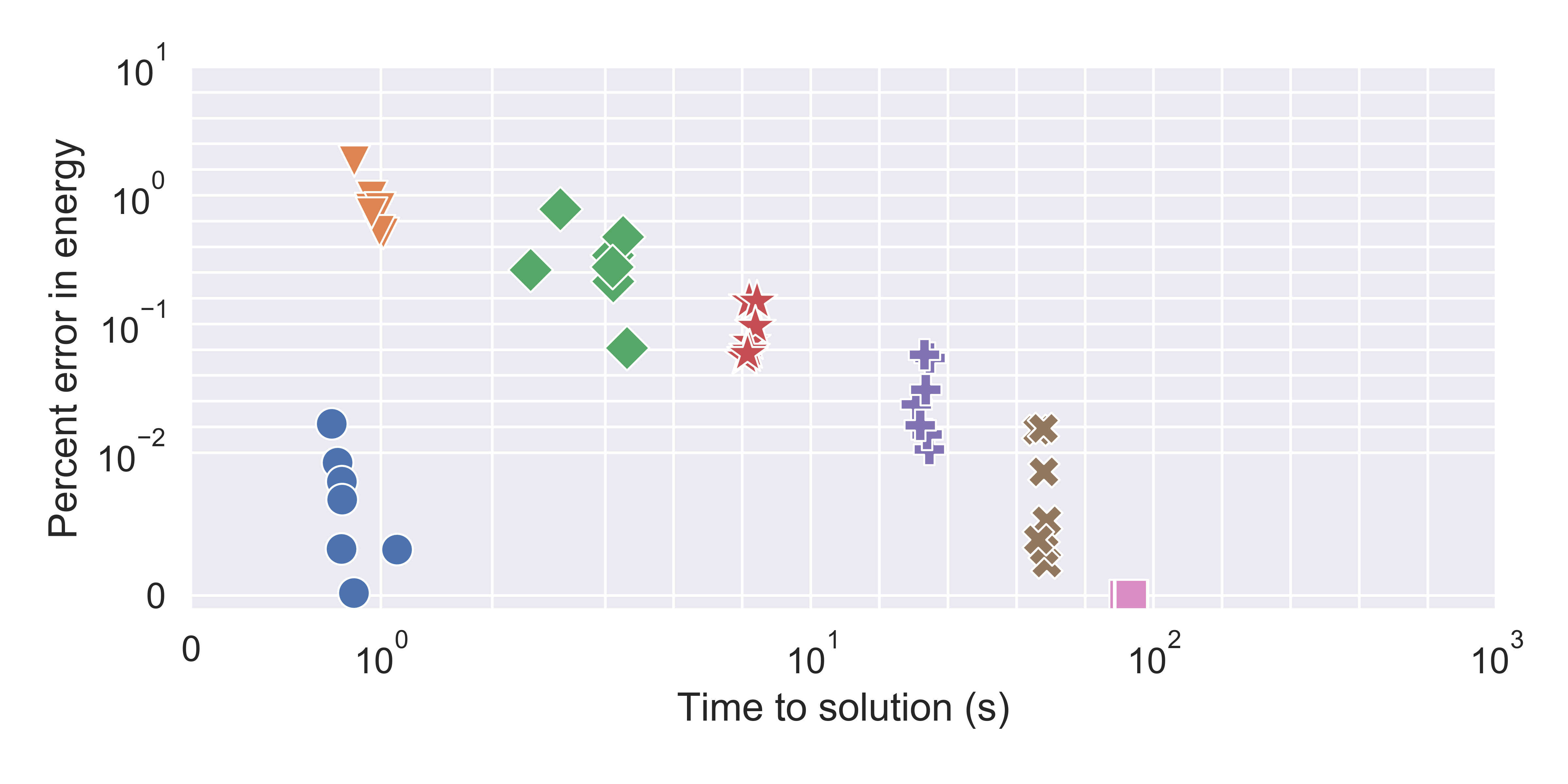}
 \end{adjustbox}
}}&%
	{\resizebox{0.45\textwidth}{!}{
 \begin{adjustbox}{clip, trim=.3cm 0cm .7cm .3cm}
	\includegraphics{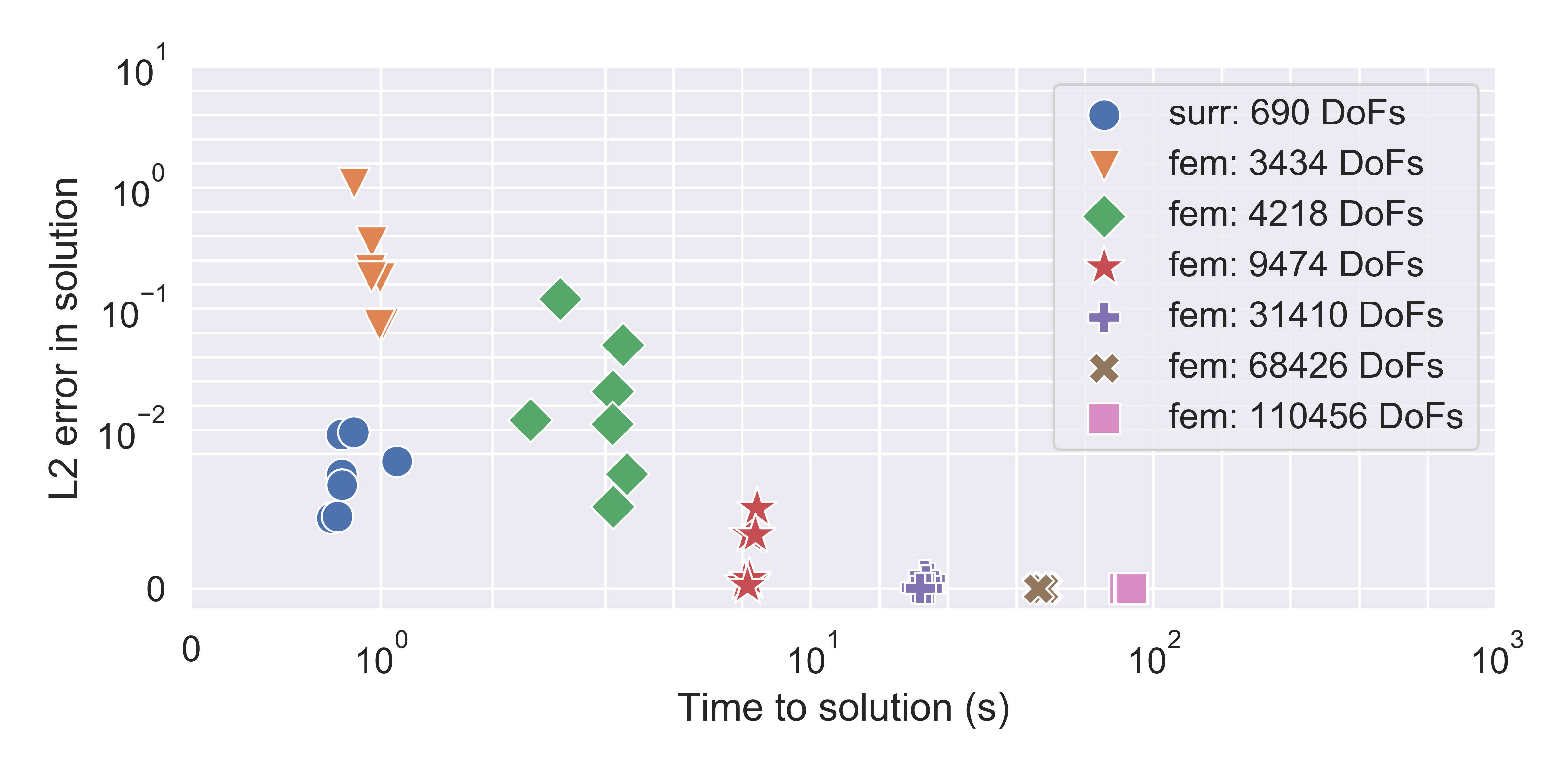}
	 \end{adjustbox}
	}}
\end{tabular}
	\vspace{-0.5cm}%
	\caption{Error in solution and in estimated energy vs solution wall clock time for the composed energy surrogate and for finite element models with varying mesh sizes. Top: axial compression. Bottom: axial tension.}%
	\label{fig:results}%
 \vspace{-0.3cm}
\end{figure*}

\section{Data and training}
\label{sec:training}
We carry out data collection in two phases.
First, we collect training and validation datasets using Hamiltonian Monte Carlo \citep{duane1987hybrid} to preferentially sample displacements which correspond to lower energy modes.
Next, we perform dataset aggregation \citep{ross2011reduction} to augment the dataset such that the learned energy model will be accurate on the states it encounters when deployed.
\vspace{-0.2cm}
\subsection{Solving the PDE}
To collect training data, we use the reduced-basis displacement $\tilde{u}$ corresponding to a vector of spline coefficients $\rvu$ as the boundary condition around a domain $\Omega$ representing a $2\times2$-pore subdomain, and solve the PDE
\begin{align*}
\text{Div } F(u) &= 0 \in \Omega \\
 u &= \tilde{u} \in \partial\Omega
\end{align*}
using a load-stepped relaxed Newton's method \citep{sheng2002automatic}. The relaxed Newton's method takes the iteration:
\begin{equation}
\vspace{-0.1cm}
\vec{u} \leftarrow \vec{u} - \lambda
(\frac{\partial^2 E}{\partial \vec{u}^2})^{-1}\frac{\partial E}{\partial \vec{u}}
\end{equation}
Above, $0 < \lambda < 1$ is the relaxation parameter (analogous to a step size), and $\vec{u}$ is the vector of coefficients defining $u$ in the FEA basis.
Newton's method requires an initial guess which is sufficiently close to the true solution \citep{kythe2004introduction}.
Smaller relaxation parameters yield a greater radius of convergence but necessitate more steps to solve the PDE.

The radius of convergence can also be aided by load-stepping: solving the PDE for a sequence of boundary conditions, annealing from an initial boundary condition for which we have a good initial guess (e.g., the rest configuration) to a final boundary condition~$\tilde{u}$, using the solution to the previous problem as an initial guess for Newton's method for the next problem.
We find that combining load stepping with a relaxed Newton's method allows problems to be solved more efficiently than using either alone.
Except where specified, we linearly anneal from rest to~$\tilde{u}$ over~$10$ load steps and use a relaxation parameter~$\lambda = 0.1$.
\vspace{-0.2cm}
\subsection{Initial dataset collection}
We wish to train on a wide variety of displacement boundary conditions. Solution procedures minimize the energy: thus, lower energy modes will be encountered in the solve. We choose a shaping distribution where the density is the product of a Boltzmann density~$\exp \{\tilde{E}\}/Z$ and a Gaussian density~$\mathcal{N}(\bar{x}(\rvu); \bar{\mu}, \Sigma)$, where~${\bar{x}(\rvu) \in \mathbb{R}^{2\times2}}$ is a macroscopic strain tensor\footnote{See the appendix for approximating $\bar{x}$ from $\rvu$.} represented by $\rvu$, $\bar{\mu}$ is a target macroscopic strain drawn from an i.i.d.\ Gaussian with standard deviation~$0.15$, and~$\Sigma$ is set to~$(\bar{\mu}\circ \bar{\mu})^{-1}$. 

Given a solution to the PDE, the log-density and its displacement may be cheaply computed (the latter via the adjoint method). Making use of these gradients, we sample data points with Hamiltonian Monte Carlo (HMC). After sampling a data point, we compute the corresponding Hessian and save the tuple $(\rvu, \xi, \tilde{E}, \nabla_{\rvu} \tilde{E}, \nabla^2_{\rvu} \tilde{E})$ as a data point.

We initialize each HMC data collector by sampling a macroscopic displacement target and a random pore shape. We do not use load-stepping, instead using the solution for the $\rvu$ used in the previous iteration of HMC's leapfrog integration as an initial guess for solving the PDE. We randomize HMC hyperparameters for each collector to attempt to minimize the impact of specific settings: see the appendix for exact ranges. We sample \num{55000} training examples and \num{5000} validation examples altogether. We visualize displacements drawn from this distribution in the appendix.
\vspace{-0.2cm}
\subsection{Data aggregation}
The procedure of deploying the surrogate defies standard i.i.d.\ assumptions in supervised machine learning.
That is, when deployed, the surrogate encounters states determined by the energy it defines and on the boundary conditions placed on the composed body.
Depending on this energy and on the boundary conditions, the surrogate may encounter states which do not resemble those sampled with HMC.

This problem---that training an agent to predict expert actions with supervised learning leads to trajectories dissimilar to those on which it was trained---is a central concern in the imitation learning literature. A number of solutions exist \citep{schroecker2017state}.
One such technique is dataset aggregation, or \textsc{DAgger} \citep{ross2011reduction}, an extension of earlier approaches SEARN \citep{daume2009search} and SMILe \citep{ross2010efficient}, which reduces imitation learning or structured prediction to online learning.

In \textsc{DAgger}, a policy is deployed and trajectories are collected.
The expert is queried for ground-truth actions on the states in these trajectories.
The state-action pairs are appended to the dataset, and the policy is retrained on this dataset.
This process of deployment, querying, appending data, and retraining, is iterated.
Under appropriate assumptions, the instantaneous regret of the learned policy vanishes with the number of iterations, i.e., the learned policy will match the expert policy on its own trajectories.

\citet{ross2011reduction} present \textsc{DAgger} as a method for discrete action spaces.
Our problem has a continuous action space: the gradient of the energy in a cell.
We do not investigate whether it is possible to generalize \textsc{DAgger}'s regret guarantees to continuous action spaces, but the intuition holds that we wish our model to ``imitate'' the finite element ``expert'' on the optimization trajectories the model produces.

We initialize our training data as described in the preceding section.
We then apply \textsc{DAgger} by iterating: (i) training the surrogate; (ii) composing surrogates and finding the displacements which minimize the composed energy; (iii) sampling displacements which lie along the surrogate's solution path, querying the ground-truth energy and energy derivatives using the finite element model, and adding these new data points to the dataset.
We visualize displacements generated by \textsc{DAgger} in the appendix.
\vspace{-0.3cm}
\section{Software, hardware, and systems}
We implement the finite element models in~\texttt{dolfin}~\citep{logg2010dolfin, logg2012dolfin}, a Python front end to FEniCS \citep{alnaes2015fenics,logg2012automated}. To differentiate through finite element solutions, we use the package~\texttt{dolfin-adjoint}~\citep{mitusch2019dolfin}. We implement surrogate models in PyTorch \citep{paszke2019pytorch}.

We use Ray \citep{moritz2018ray} to run distributed workloads on Amazon EC2. The initial dataset is collected using 80 M4.xlarge CPU spot workers. While training the surrogate, we use a GPU P3.large driver node to train the model, and 80 M4.xlarge CPU spot worker nodes performing \textsc{DAgger} in parallel. These workers receive updated surrogate model parameters, compose and deploy the surrogate, sample displacements along the solution path, query the finite element model for energy and derivatives, and return data to the driver node. Initial dataset collection and model training with \textsc{DAgger} each take about one day in wall-clock time.
\vspace{-0.3cm}

\section{Empirical evaluation}
\label{sec:results}
In this section we demonstrate the ability of Composable Energy Surrogates (CES) to efficiently produce accurate solutions. We compare wall-clock computation time and solution accuracy of CES to that of FEA at varying fidelities.

We consider the systems constructed in \citet{overvelde2014relating}: structures with an $8\times8$ array of pores, corresponding to a $4\times4$ assembly of our surrogates, each of which represents a $2\times2$-pore component. We sample pore shapes from a uniform distribution over the valid shapes defined in \citet{overvelde2014relating}. For \textsc{DAgger}, we sample macroscopic vertical axial strain magnitudes from $\mathcal{U}(0., 0.3)$, and choose to apply compression with probability $0.8$ (as compressive displacements involve more interesting pore collapse) or tension with probability $0.2$.

We compare our composed surrogates to finite element analysis with a range of different-fidelity meshes under axial compression and tension with a macroscopic displacement of~$0.125 L_0$, where~$L_0$ is the original length of the solid. See the appendix for details of the finite element meshes. Comparison is carried out for seven pore shapes: $\xi=(0, 0)$, corresponding to circular pores, and six $\xi$ sampled from a uniform distribution over pore parameters defined as valid in \citet{overvelde2014relating} via rejection sampling.

For the composed surrogate, we use PyTorch's L-BFGS routine to minimize the energy, with fixed step size $0.25$ and PyTorch's default criteria for checking convergence. We attempt to solve each finite element model with Newton's method with load steps $[1, 2, 5, 10, 20]$ and relaxation parameters $[0.9, 0.7, 0.4, 0.1, 0.05]$. We record the time taken for the \emph{fastest} solve which converges. Under compression these solids exhibit nonlinear behavior, and only the more conservative solves converge. Under tension they behave closer to a linear elastic model, and Newton's method converges quickly. Measurements are taken on an Amazon AWS M4.xlarge EC2 CPU instance. Using a GPU could provide further acceleration for the composed surrogate.

We measure error in the solution and in the macroscopic energy. The former is~${||\hat{u}-u^*||_2^2}$, where $\hat{u}$ and $u^*$ are the approximation and ground-truth evaluated at the spline control points. We also measure the relative error, $\nicefrac{|\hat{E}(\hat{u}) - E^*(u^*)|}{E^*(u^*)}$, where $\hat{E}(\hat{u})$ is the approximated energy of the approximate solution, and $E^*(u^*)$ is the ground-truth energy of the ground-truth solution. As the energy function determines macroscopic behavior, accuracy of this energy is a potential indicator of a model's ability to generalize to larger structures. The highest-fidelity finite element model is used for the ground truth $E^*$ and $u^*$, and has an error of zero on both metrics by definition. Multiple solutions exist as energy is preserved under vertical and horizontal flips, so before comparing a solution~$\hat{u}$ to the ground-truth~$u^*$ we programatically check these flips and use the flip which minimizes the Euclidean distance.

Figure \ref{fig:results} shows our evaluation. Composed energy surrogates are more efficient than high-fidelity FEA simulations yet more accurate than low-fidelity simulations. CES produces solutions with equivalent~$\ell^2$ error to FEA solutions which need an order of magnitude more variables and computation time, or produces solutions with an order of magnitude less~$\ell^2$ error than FEM solutions requiring the same computation. This gap increases to several orders of magnitude when we consider percentage error in the predicted strain energy. We visualize the ground-truth and the CES approximation in Figure \ref{fig:compress}. See the appendix for visualization of the FEM solutions and of CES for the remaining structures.
\vspace{-0.2cm}
\begin{figure}[h]
 \begin{tabular}{c|c}
	 {\resizebox{0.40\linewidth}{!}{
 \begin{adjustbox}{clip, trim=3.35cm 1.7cm 2.7cm 1.8cm}
	 \includegraphics{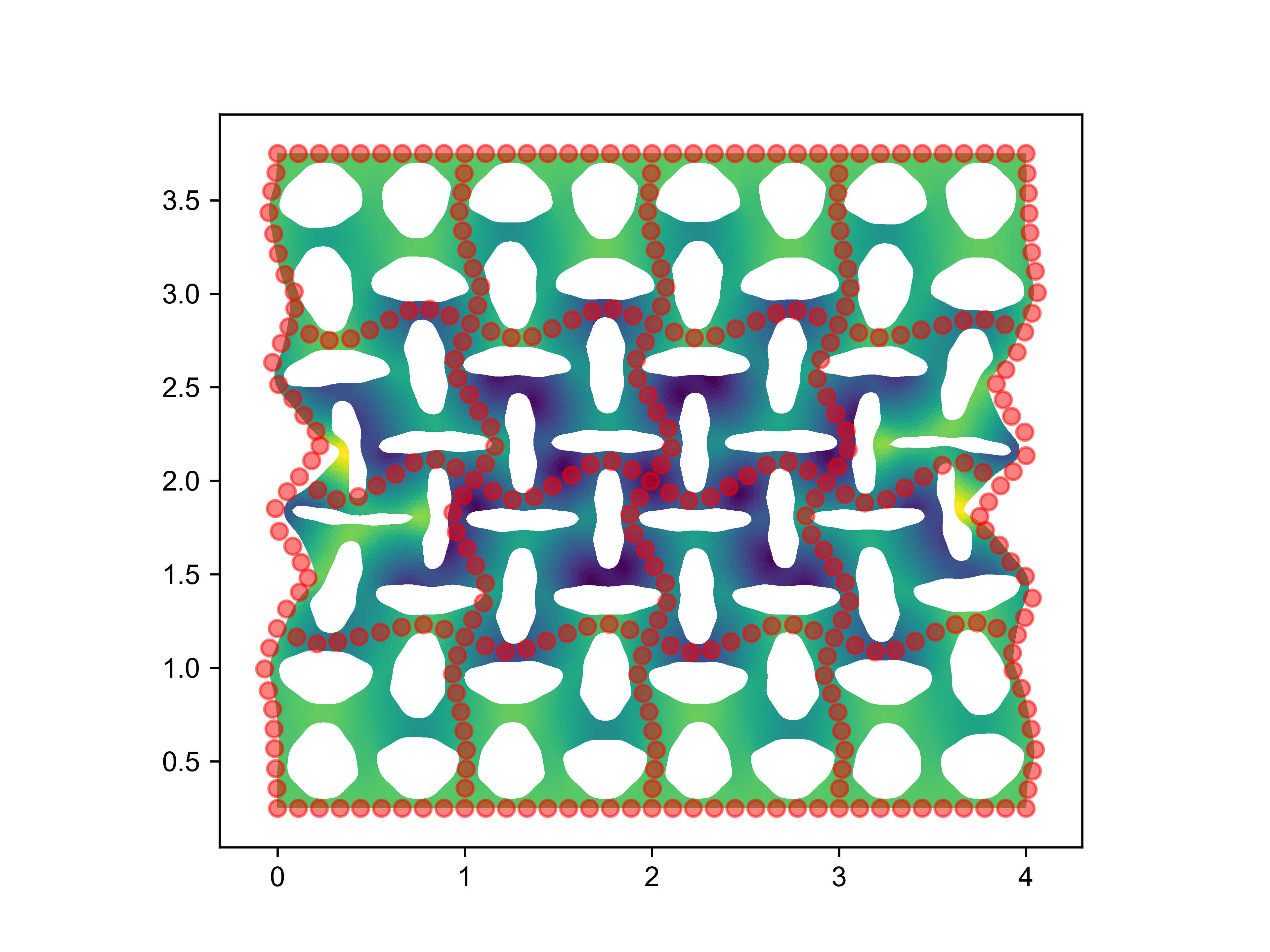}
	 \end{adjustbox}
	 }}&
	{\resizebox{0.45\linewidth}{!}{
 \begin{adjustbox}{clip, trim=2.2cm 1.7cm 2.3cm 1.8cm}
	\includegraphics{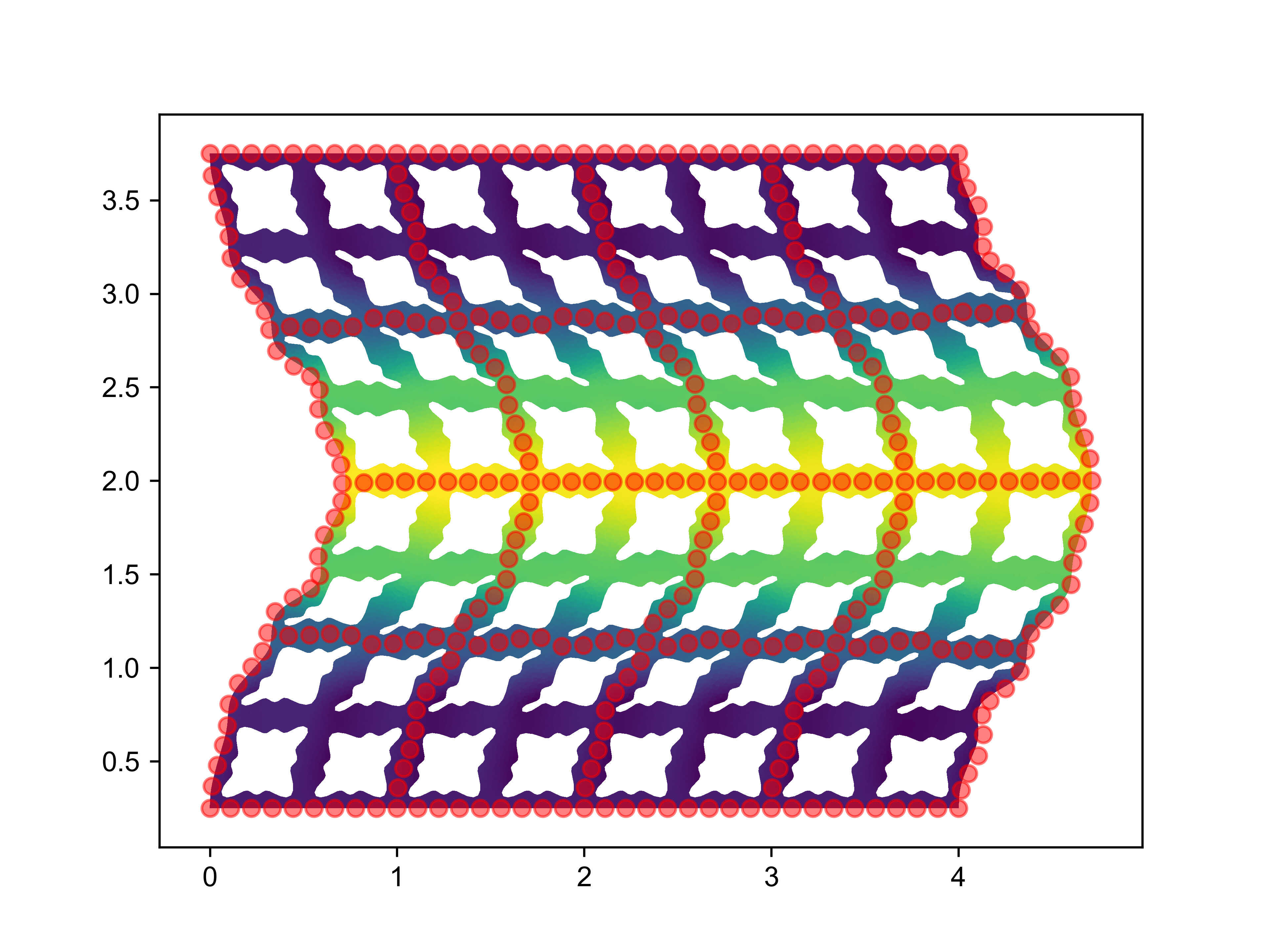}
	 \end{adjustbox}
	}}\\\hline
		\rule{0pt}{13ex}
	 {\resizebox{0.37\linewidth}{!}{
 \begin{adjustbox}{clip, trim=4.5cm 1.5cm 4.0cm 1.5cm}
	 \includegraphics{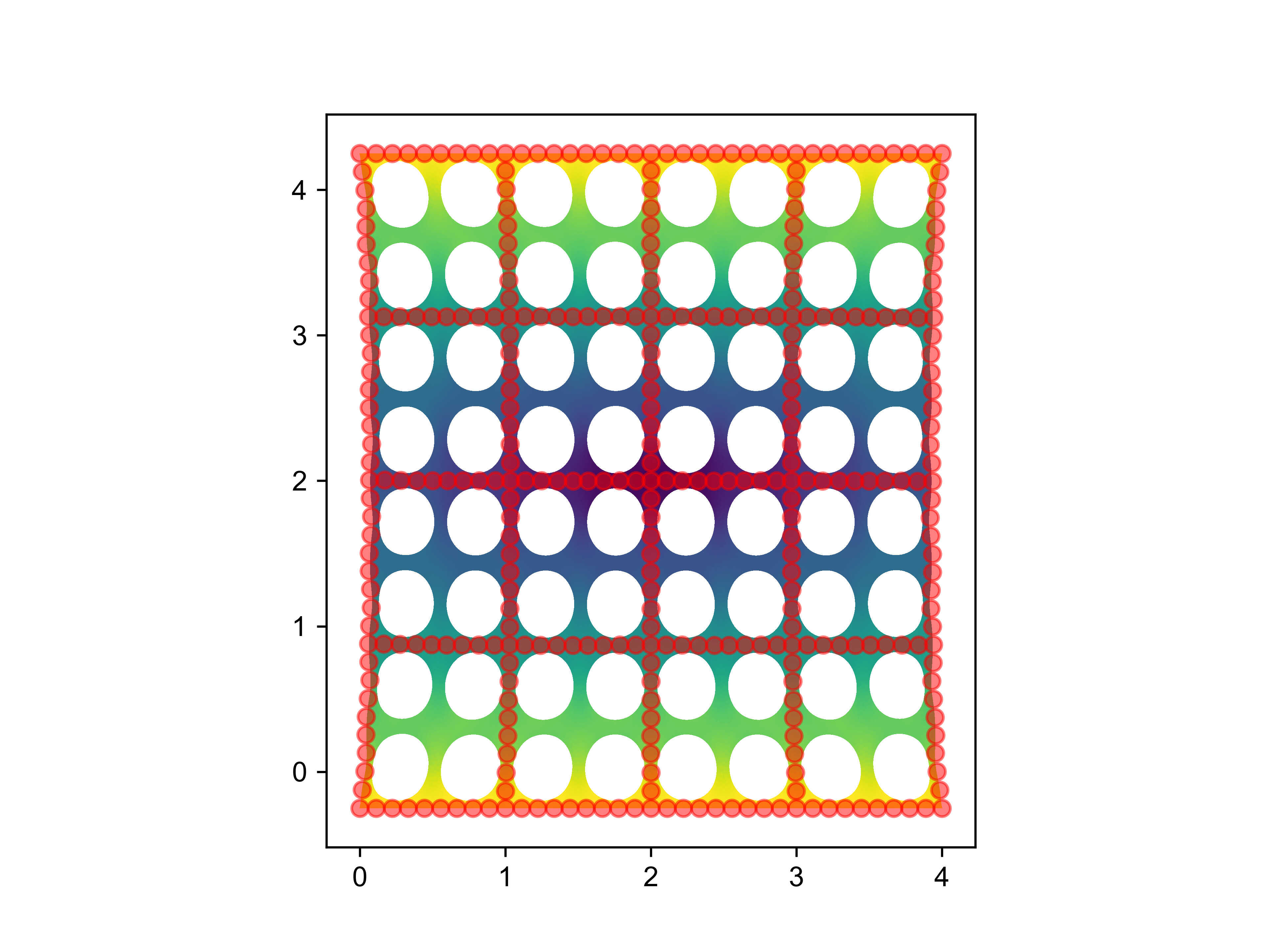}
	 \end{adjustbox}
	 }}&
	{\resizebox{0.37\linewidth}{!}{
 \begin{adjustbox}{clip, trim=4.5cm 1.5cm 4.0cm 1.5cm}
	\includegraphics{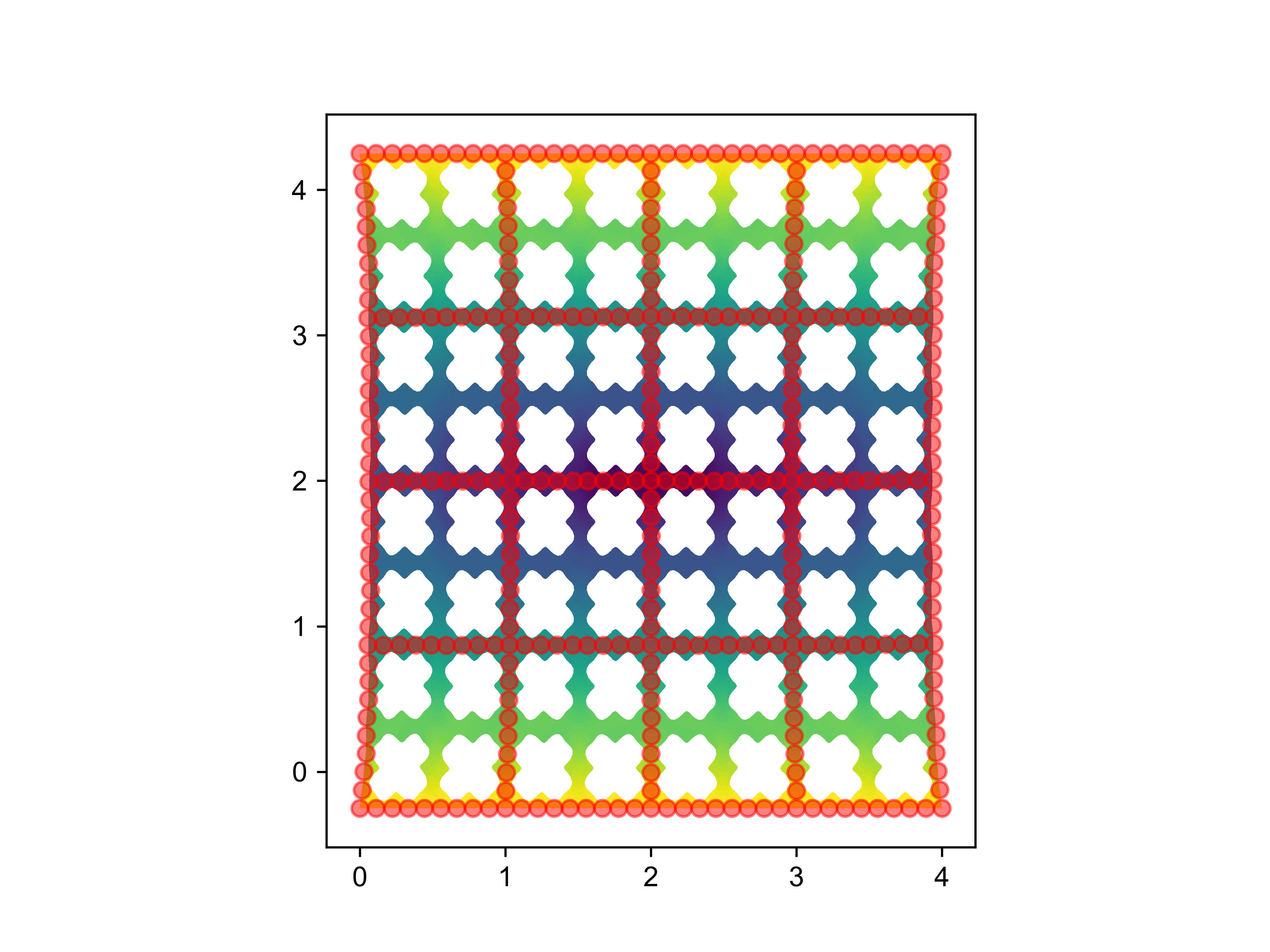}
	 \end{adjustbox}
	}}
\end{tabular}
\vspace{-0.2cm}
\caption{Meta-materials under axial compression (top) and tension (bottom), with the solution found via CES shown in red at the spline control points. The CES-approximated solution approximates the FEA solution to high visual fidelity.}
\vspace{-0.8cm}
\label{fig:compress}
\end{figure}
\vspace{-0.4cm}

\section{Limitations and opportunities}
\label{sec:discussion}
\textbf{Use of \textsc{DAgger}}. We use \textsc{DAgger} to assist CES to match the ground-truth on the states encountered during the solution trajectory. \textsc{DAgger} requires one to specify in advance the conditions under which the surrogate will be deployed, limiting the surrogate's use for arbitrary downstream tasks. Investigating CES's ability to generalize to novel deployment conditions--and designing surrogates which can do so effectively--is an important direction for future work.

\textbf{Error estimation, refinement, and guarantees}. Finite element methods enjoy advantageous properties. As one uses a higher-resolution mesh or higher-order elements, the solution in the finite element basis approaches the true solution. This provides a straightforward way to estimate the error (compare to the solution in a more-refined basis) and control it (via refinement). CES currently lacks these properties.

\textbf{Finite element baseline}. There is an immense body of work on finite element methods and iterative solvers. We try to provide a representative baseline, but our work should not be taken as a comparison with the ``state-of-the-art''. We aim to show that composable machine-learned energy surrogates enjoy some advantages over a reasonable baseline, and hold promise for scalable amortization of solving modular PDEs.

\textbf{Hyperparameters}. Both our method and the finite element baseline rely on a multitude of hyperparameters: the size of the spline reduced basis; the size and learning rate of the neural network; the size and degree of the finite element approximation; and the specific variant of Newton's method to solve the finite element model. We do not attempt a formal, exhaustive search over these parameters.

\textbf{Known structure}. We leave much fruit on the vine in terms of engineering structure known from the true equations into our surrogate. For example, we do not engineer invariance of the energy to flips and rotations of the spline coefficients. One could also use a more expressive normalizer than $||\rvu||_2^2$, e.g. the energy predicted by a coarse-grained linear elastic model, or exploit spatially local correlation, e.g. by using a 1-d convolutional network around the boundary of the cell.
\vspace{-0.2cm}

\section{Conclusion}
We present a framework for collapsing optimization problems with a local bilevel structure by learning composable energy surrogates. This framework is applied to amortizing the solution of PDEs corresponding to mechanical meta-material behavior. Learned composable energy surrogates are more efficient than high-fidelity FEA yet more accurate than low-fidelity FEA, occupying a new point on the Pareto frontier. We believe that learning composable energy surrogates could accelerate metamaterial design, as well as design and identification of other systems described by PDEs with parametric modular structure.

\section{Acknowledgements}
We would like to thank Alexander Niewiarowski for numerous helpful discussions about continuum mechanics and FEA, Ari Seff for help finding a particularly difficult bug in our data pipeline, and Maurizio Chiaramonte for inspiring early conversations about metamaterials and model order reduction. This work was funded by NSF
IIS-1421780 and by the Princeton Catalysis Initiative.

\bibliographystyle{icml2020}
\bibliography{references}

\appendix
\onecolumn
\end{document}


\begin{frontmatter}
\title{Appendix}

\begin{aug}
\author{}%

\runauthor{Anon}

\address{}
\end{aug}

\begin{abstract}
This appendix consists of:
(i) specification of the data generating distribution and hyperparameters;
(ii) visualization of data generated via Hamiltonian Monte Carlo;
(iii) visualization of data generated via \textsc{DAgger};
(iv) hyperparameters used for neural network specification and training;
(v) ablation study of neural network design choices;
(vi) specification of the finite element meshes used as baselines;
(vii) visualization of all solutions found under compression and tension
for each pore shape for each baseline mesh and the composed energy surrogate.
\end{abstract}

\end{frontmatter}

\section{Data generation with Hamiltonian Monte Carlo}
We use 100 data collectors, each with randomly drawn hyperparameters, which
each terminate (and have their place taken by a newly initialized collector)
after sampling 25 data points.
We collect 60,000 data points, consisting of a training set of 55,000 and a validation set of 5,000.
As our distribution is arbitrary, and as we assume that more data is always a good thing, when a HMC sample is rejected, we still add it to the dataset, but return to the last un-rejected sample to continue the Markov chain.

Hyperparameter distributions are chosen heuristically such that the finite element simulation tends to converge in a reasonable amount of time.
The hyperparameter distributions are as follows:

\begin{itemize}
\item Leapfrog step size: $\mathcal{U}(0.005, 0.02)$
\item Leapfrog path length: $\mathcal{U}(0.05, 0.3)$
\item Temperature used to scale the log-probability: $\mathcal{U}([0.0001,0.0005,0.001,0.005,0.01,0.05,0.1])$
\item Standard deviation of the Gaussian from which the Hamiltonian momentum is drawn: $\mathcal{U}(0.01, 0.3)$
\end{itemize}

We approximate the macroscopic strain tensor $\bar{x}$ from $\mathbf{u}$ as:

\resizebox{\linewidth}{!}{
  \begin{minipage}{\linewidth}
\begin{align*}
\bar{x}(\mathbf{u}) = \frac{1}{N} \begin{bmatrix}
\displaystyle\sum_{X \in \text{rhs}} \!\!u_1(X) \!-\!\!\! \sum_{X \in \text{lhs}} \!\!u_1(x)
&
\displaystyle\sum_{X \in \text{rhs}} \!\!u_2(X) \!-\!\!\! \sum_{X \in \text{lhs}} \!\!u_2(X)
\\
\displaystyle\sum_{X \in \text{top}} \!\!u_1(X) \!-\!\!\! \sum_{X \in \text{bot}} \!\!u_1(x)
&
\displaystyle\sum_{X \in \text{top}} \!\!u_2(X) \!-\!\!\! \sum_{X \in \text{bot}} \!\!u_2(X)
\end{bmatrix}
\end{align*}
\end{minipage}
}

Above,~$u_1(X)$ and~$u_2(X)$ are horizontal and vertical displacements defined by $\mathbf{u}$ at a point $X$, and top, bot, lhs and rhs are the set of control point locations for the splines on the top, bottom, left and right of the component.

\clearpage

\section{Visualizing HMC data}
Here we display 24 randomly chosen examples from the training set.
\begin{figure}[H]
  \centering
  \includegraphics[width=.7\linewidth]{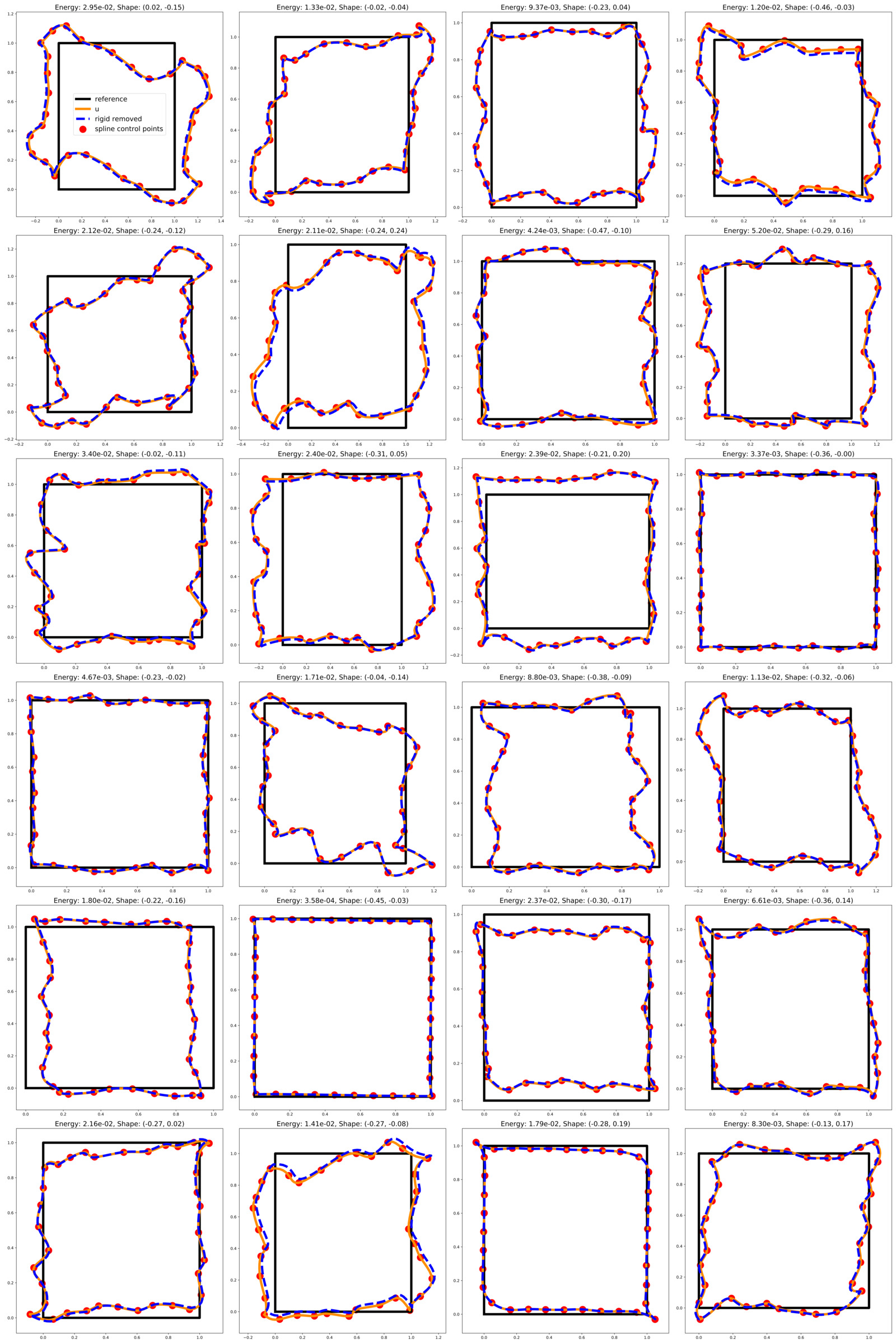}
\end{figure}

\clearpage

\section{Visualizing \textsc{DAgger} data}
Here we display 24 randomly chosen examples from the data collected with \textsc{DAgger}.
\begin{figure}[H]
  \centering
  \includegraphics[width=.7\linewidth]{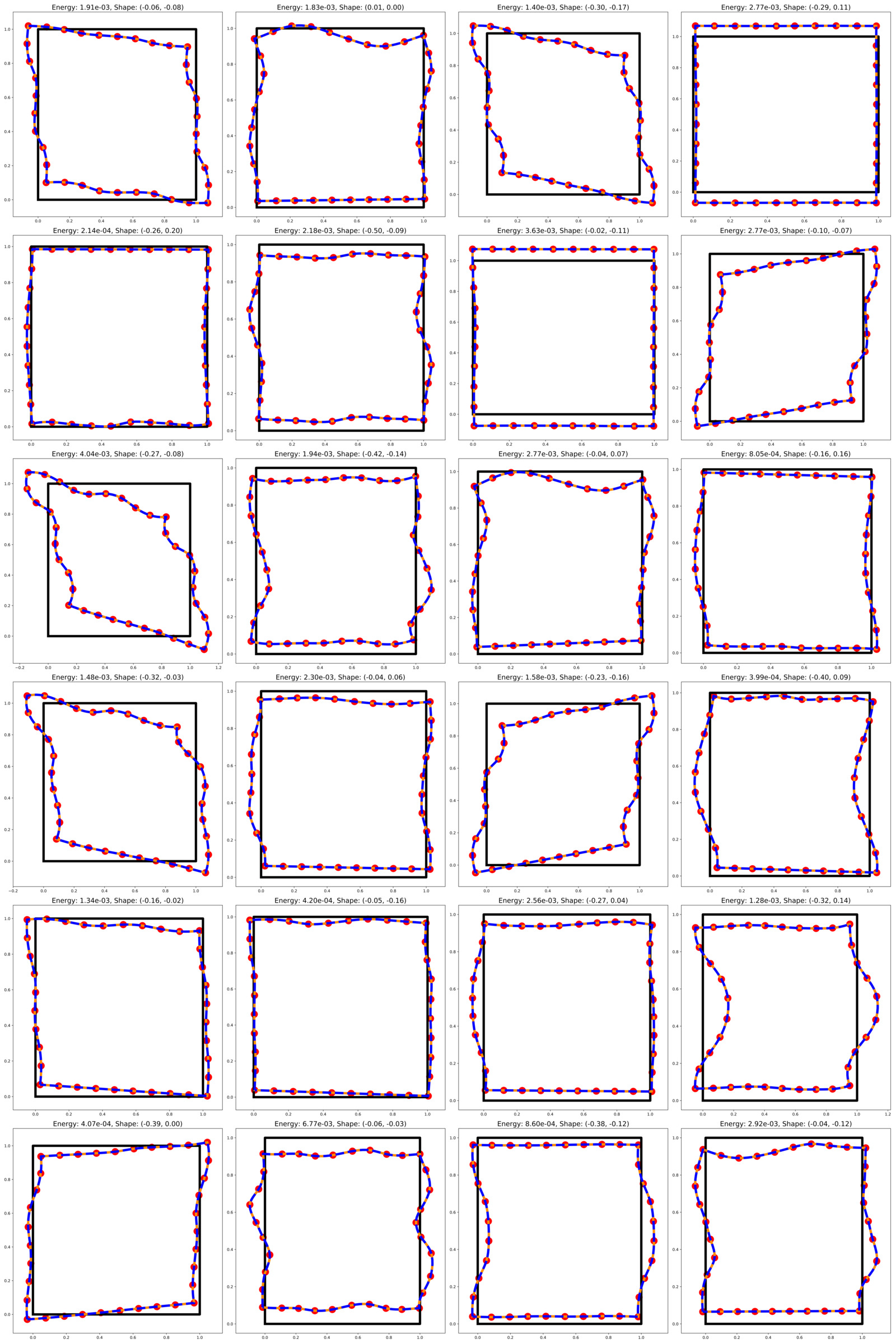}
\end{figure}

\section{Neural network hyperparameters}
We use a fully-connected neural network with three hidden layers of 512 units, Swish nonlinarities, and He initialization.
We optimize our neural network using Adam with a learning rate of 3e-4 and a batch size of 512.

\section{Surrogate design ablation study}
We perform an ablation study by switching on and off the following independent variables:
\begin{itemize}
\item "Scale": parameterizing the log of the scalar stiffness, vs parameterizing energy directly;
\item "Remove rigid": removing rigid body translations from the data via Procrustes analysis;
\item "Sobolev-G": Sobolev training on energy gradients;
\item "Sobolev-Hvp": Sobolev training on energy Hessian-vector products.
\end{itemize}
We measure performance after 10,000 training steps (93 epochs) on the training set, without \textsc{DAgger}.
We evaluate each model on the validation dataset using the following metrics:
\begin{itemize}
\item "E \%err": the error in predicted energy, expressed as a percentage of the true energy;
\item "G-sim": the cosine similarity between predicted and true gradients;
\item "Hvp-sim": the cosine similarity between predicted and true Hessian-vector products.
\end{itemize}
Results are shown below. For the independent variables, a value of '1' indicates that method or technique was turned on, while a value of '0' indicates it was turned off.

Each design choice improves the validation metrics.
"Remove rigid" has marginal impact, as our training displacements contain little rigid body transformation.
We leave this feature in as it causes no harm;
as it improved performance under earlier dataset creation methods which resulted in more rigid body translation;
and as removing translations before computing energy is necessary to be able to compose energy surrogates by tiling.

\begin{figure}[H]
  \centering
  \includegraphics[width=.8\linewidth]{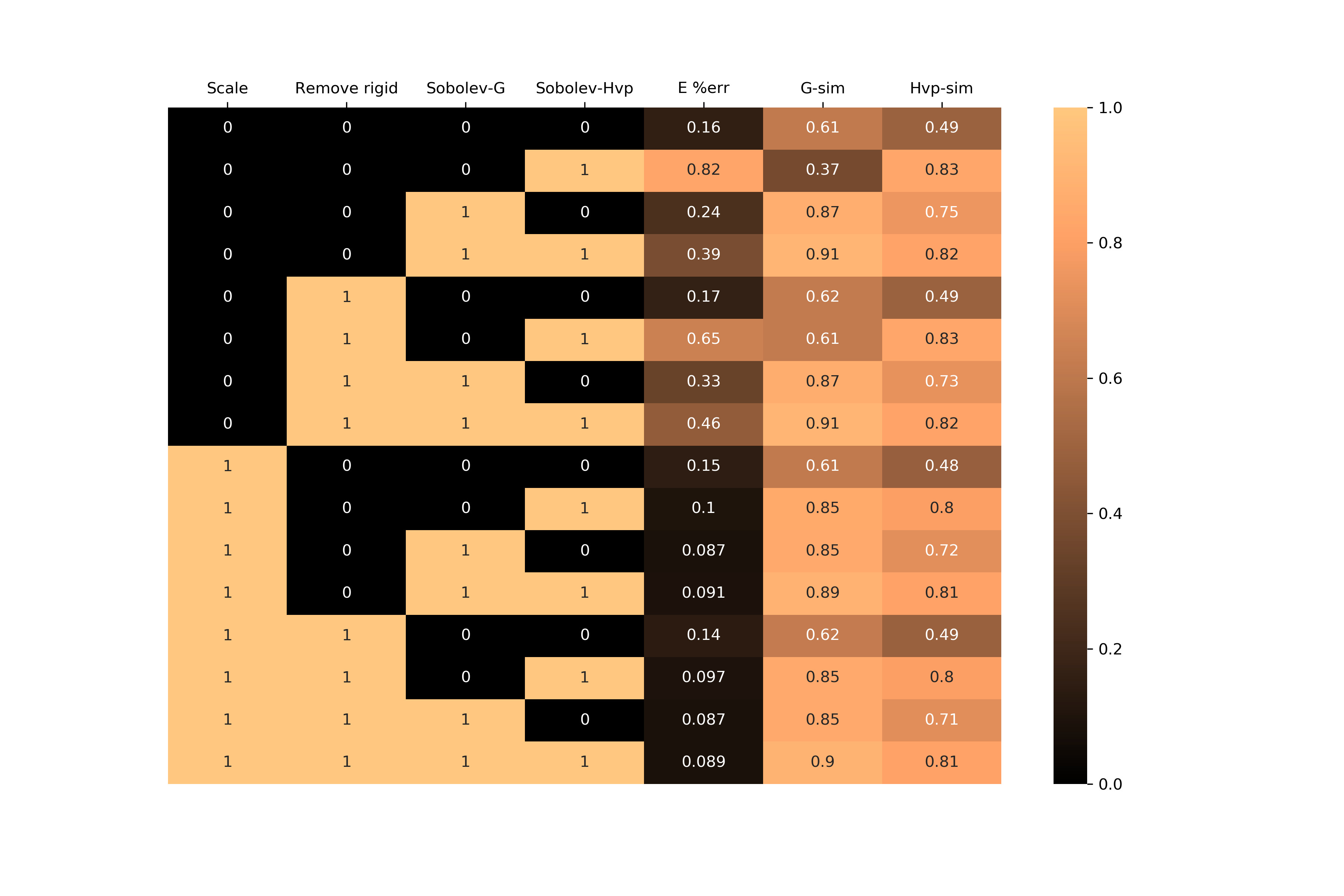}
\end{figure}

\section{Finite element baselines}
We generate meshes for the finite element baselines using two parameters: pore resolution, and minimum mesh resolution.
Firstly, for each pore in the cellular solid, we generate a polygon representing that pore using a number of points equal to pore resolution.
We let the material domain be the overall volume of the solid with these polygons subtracted.
Next, we generate a mesh over the material domain using \textsc{mshr}'s automated mesh generation routine,
passing as resolution minimum mesh resolution multiplied by the number of cells.
In \textsc{mshr}, the resolution parameter controls the maximum cell size, which
is the diameter of the domain's bounding circle divided by the resolution.
It should be noted that cells can be much smaller than this maximum size, or there
can be many more cells than the resolution parameter would imply, as \textsc{mshr}
will place one cell vertex on each point used to construct the domain geometry (i.e. each point in the pore polygon).

\section{Benchmark visualizations}
In the following pages we visualize the solutions found for each pore by each FEA mesh and by CES.
For each pore we use six different finite element meshes.
These respectively used $[4, 8, 16, 32, 48, 64]$ points used to define the geometry of each pore, and minimum of $[1, 2, 4, 8, 12, 16]$ internal mesh vertices along a given axis per pore.
Given these parameters and the geometry of the material domain, meshes were created using the automatic mesh generation tool from mshr (the mesh generation component of FEniCS).
We include the number of degrees of freedom in the finite element basis in each plot.
We superimpose the solution found with CES in red dots on the solution found with FEA.
The CES solution has 690 degrees of freedom in all cases.

\clearpage

\subsection{Compression}
\begin{figure}[H]
\begin{subfigure}{.45\textwidth}
  \centering
  \includegraphics[width=.8\linewidth]{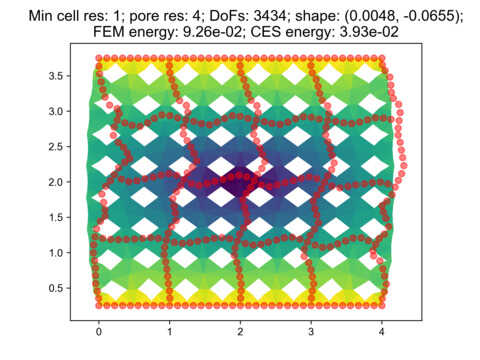}
\end{subfigure}
\begin{subfigure}{.45\textwidth}
  \centering
  \includegraphics[width=.8\linewidth]{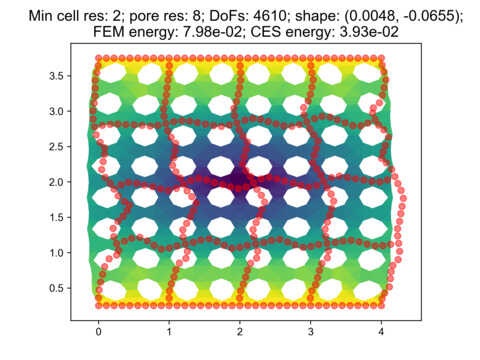}
\end{subfigure}
\newline
\begin{subfigure}{.45\textwidth}
  \centering
  \includegraphics[width=.8\linewidth]{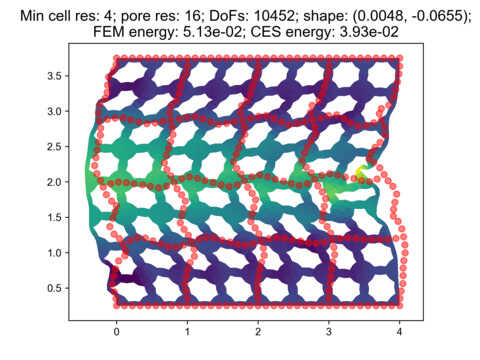}
\end{subfigure}
\begin{subfigure}{.45\textwidth}
  \centering
  \includegraphics[width=.8\linewidth]{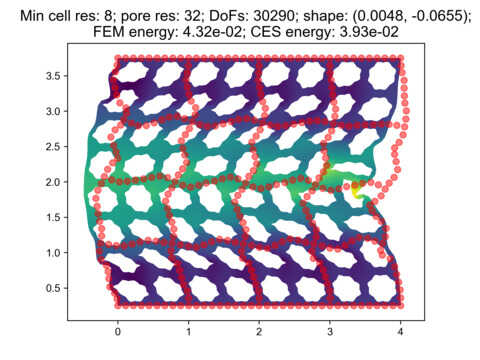}
\end{subfigure}
\newline
\begin{subfigure}{.45\textwidth}
  \centering
  \includegraphics[width=.8\linewidth]{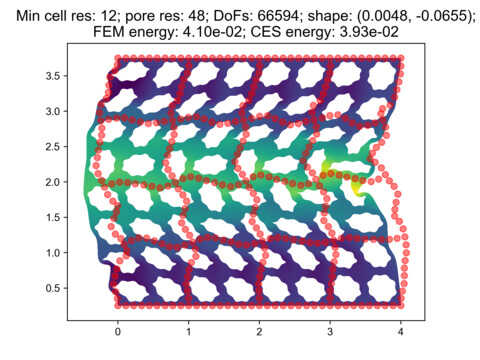}
\end{subfigure}
\begin{subfigure}{.45\textwidth}
  \centering
  \includegraphics[width=.8\linewidth]{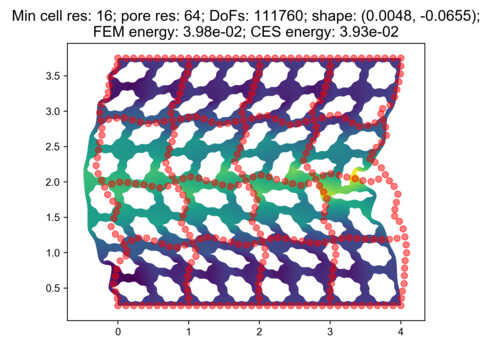}
\end{subfigure}
\end{figure}

\begin{figure}[H]
\begin{subfigure}{.45\textwidth}
  \centering
  \includegraphics[width=.8\linewidth]{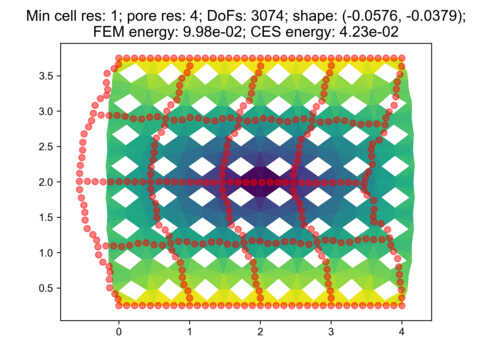}
\end{subfigure}
\begin{subfigure}{.45\textwidth}
  \centering
  \includegraphics[width=.8\linewidth]{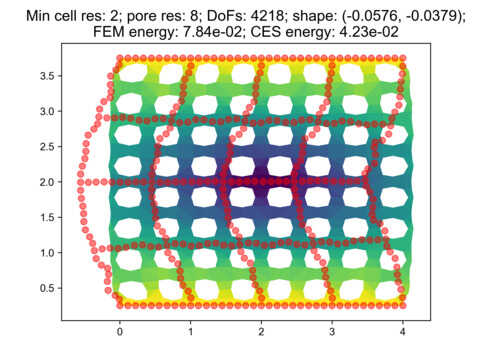}
\end{subfigure}
\newline
\begin{subfigure}{.45\textwidth}
  \centering
  \includegraphics[width=.8\linewidth]{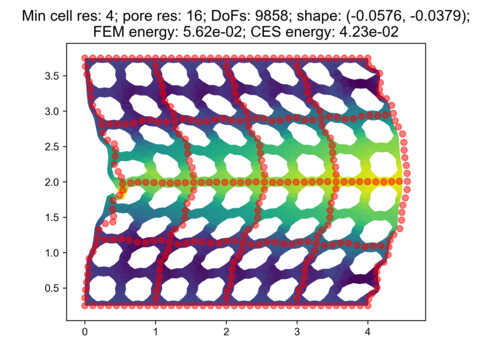}
\end{subfigure}
\begin{subfigure}{.45\textwidth}
  \centering
  \includegraphics[width=.8\linewidth]{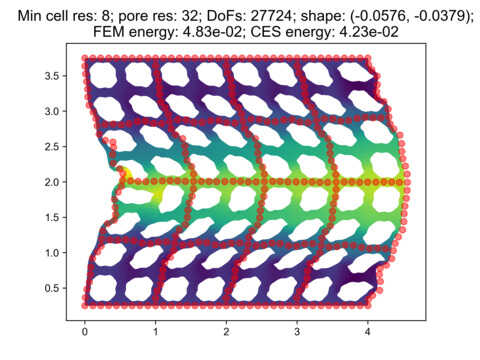}
\end{subfigure}
\newline
\begin{subfigure}{.45\textwidth}
  \centering
  \includegraphics[width=.8\linewidth]{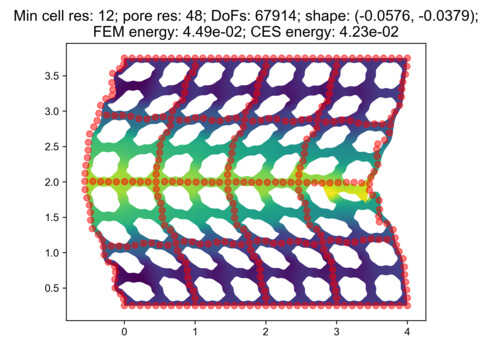}
\end{subfigure}
\begin{subfigure}{.45\textwidth}
  \centering
  \includegraphics[width=.8\linewidth]{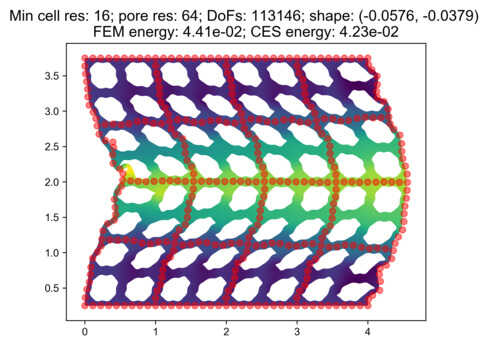}
\end{subfigure}
\end{figure}

\begin{figure}[H]
\begin{subfigure}{.45\textwidth}
  \centering
  \includegraphics[width=.8\linewidth]{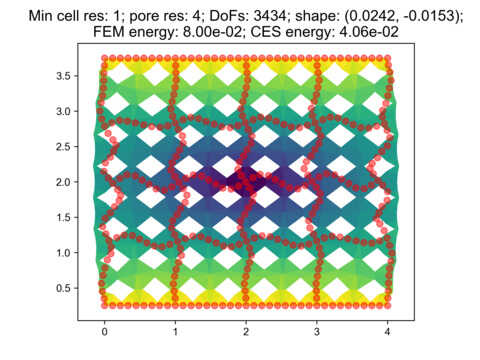}
\end{subfigure}
\begin{subfigure}{.45\textwidth}
  \centering
  \includegraphics[width=.8\linewidth]{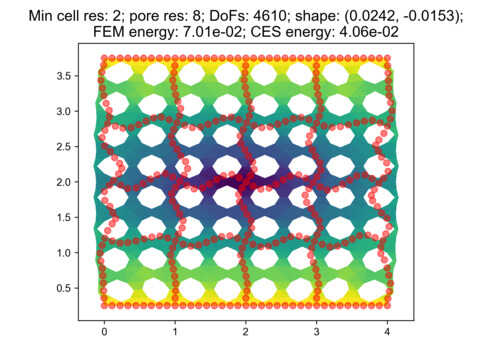}
\end{subfigure}
\newline
\begin{subfigure}{.45\textwidth}
  \centering
  \includegraphics[width=.8\linewidth]{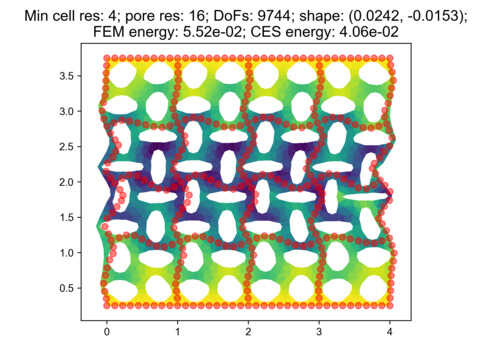}
\end{subfigure}
\begin{subfigure}{.45\textwidth}
  \centering
  \includegraphics[width=.8\linewidth]{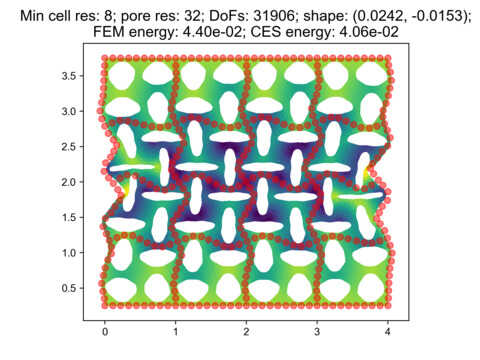}
\end{subfigure}
\newline
\begin{subfigure}{.45\textwidth}
  \centering
  \includegraphics[width=.8\linewidth]{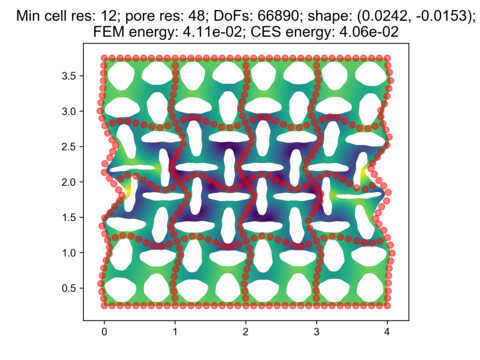}
\end{subfigure}
\begin{subfigure}{.45\textwidth}
  \centering
  \includegraphics[width=.8\linewidth]{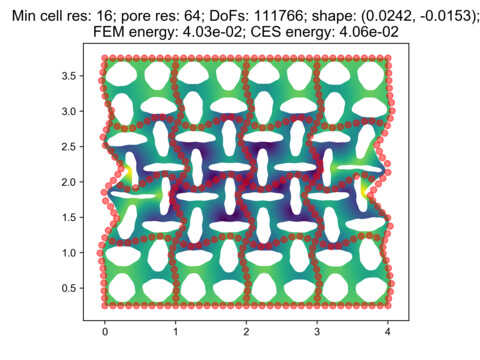}
\end{subfigure}
\end{figure}

\begin{figure}[H]
\begin{subfigure}{.45\textwidth}
  \centering
  \includegraphics[width=.8\linewidth]{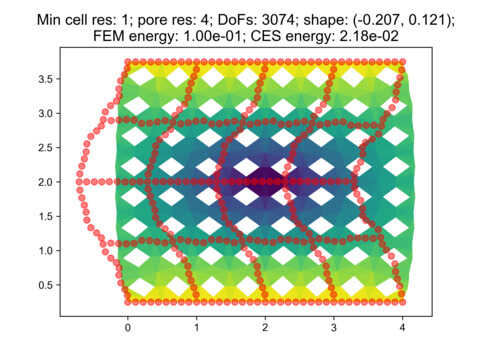}
\end{subfigure}
\begin{subfigure}{.45\textwidth}
  \centering
  \includegraphics[width=.8\linewidth]{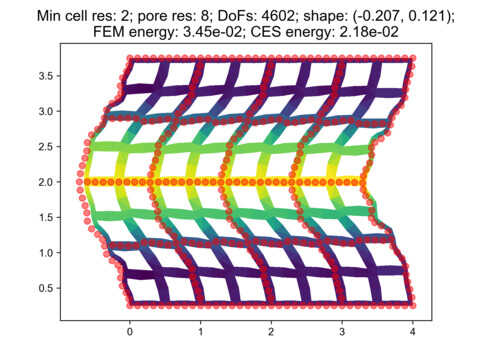}
\end{subfigure}
\newline
\begin{subfigure}{.45\textwidth}
  \centering
  \includegraphics[width=.8\linewidth]{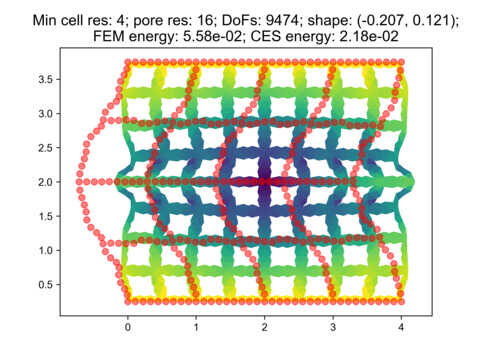}
\end{subfigure}
\begin{subfigure}{.45\textwidth}
  \centering
  \includegraphics[width=.8\linewidth]{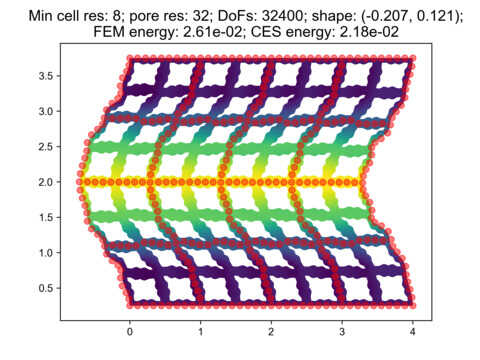}
\end{subfigure}
\newline
\begin{subfigure}{.45\textwidth}
  \centering
  \includegraphics[width=.8\linewidth]{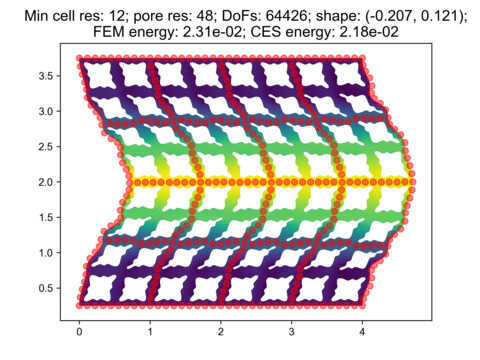}
\end{subfigure}
\begin{subfigure}{.45\textwidth}
  \centering
  \includegraphics[width=.8\linewidth]{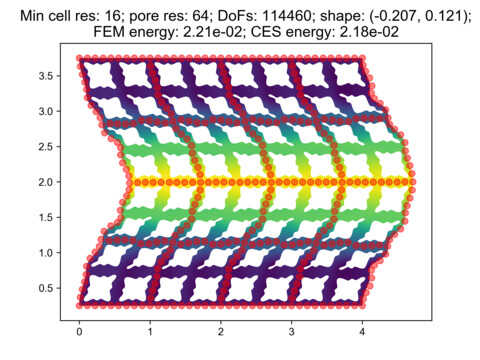}
\end{subfigure}
\end{figure}

\begin{figure}[H]
\begin{subfigure}{.45\textwidth}
  \centering
  \includegraphics[width=.8\linewidth]{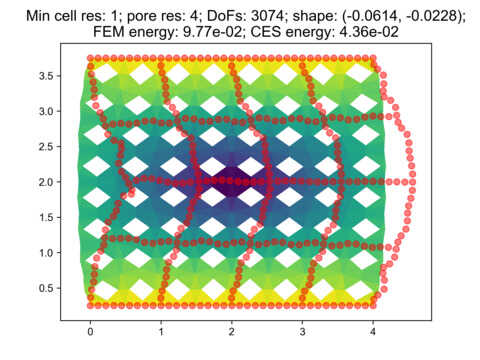}
\end{subfigure}
\begin{subfigure}{.45\textwidth}
  \centering
  \includegraphics[width=.8\linewidth]{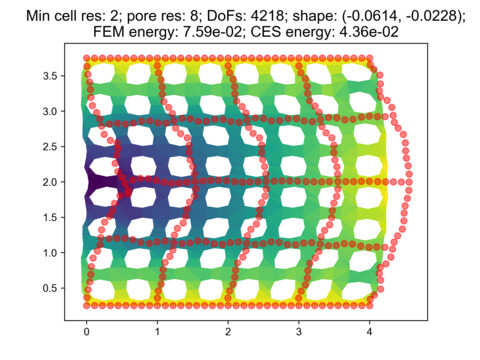}
\end{subfigure}
\newline
\begin{subfigure}{.45\textwidth}
  \centering
  \includegraphics[width=.8\linewidth]{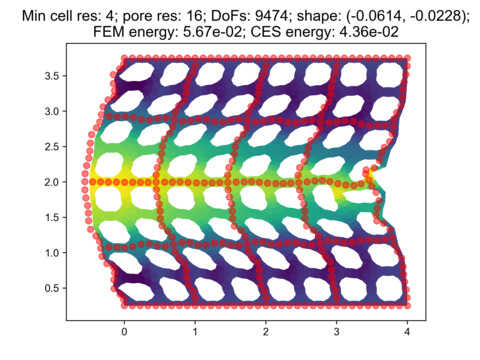}
\end{subfigure}
\begin{subfigure}{.45\textwidth}
  \centering
  \includegraphics[width=.8\linewidth]{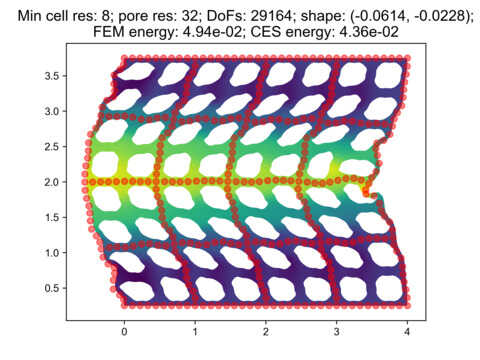}
\end{subfigure}
\newline
\begin{subfigure}{.45\textwidth}
  \centering
  \includegraphics[width=.8\linewidth]{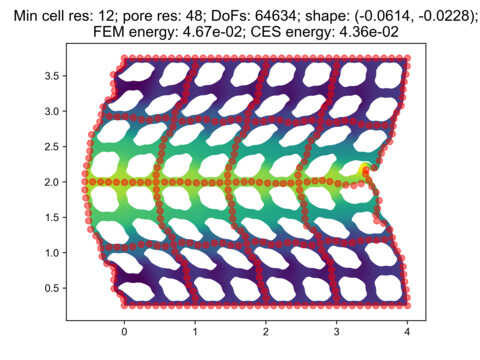}
\end{subfigure}
\begin{subfigure}{.45\textwidth}
  \centering
  \includegraphics[width=.8\linewidth]{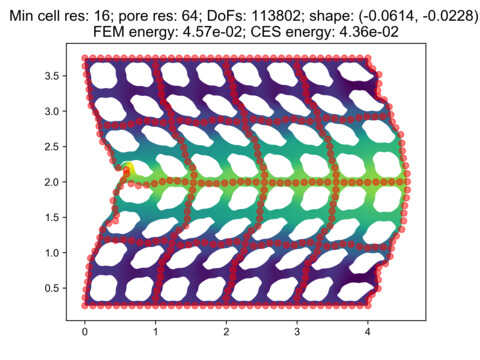}
\end{subfigure}
\end{figure}

\begin{figure}[H]
\begin{subfigure}{.45\textwidth}
  \centering
  \includegraphics[width=.8\linewidth]{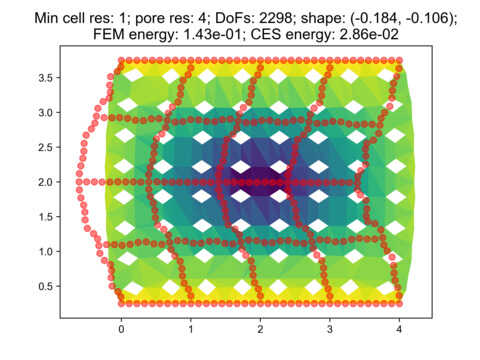}
\end{subfigure}
\begin{subfigure}{.45\textwidth}
  \centering
  \includegraphics[width=.8\linewidth]{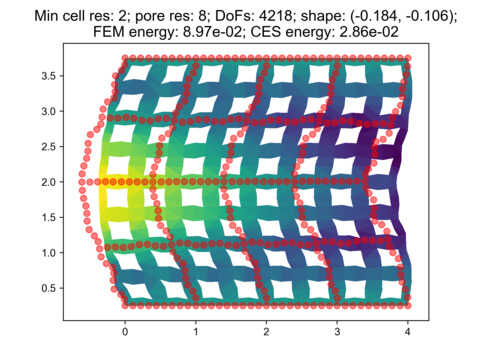}
\end{subfigure}
\newline
\begin{subfigure}{.45\textwidth}
  \centering
  \includegraphics[width=.8\linewidth]{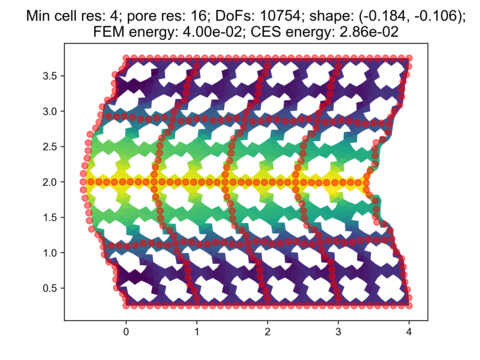}
\end{subfigure}
\begin{subfigure}{.45\textwidth}
  \centering
  \includegraphics[width=.8\linewidth]{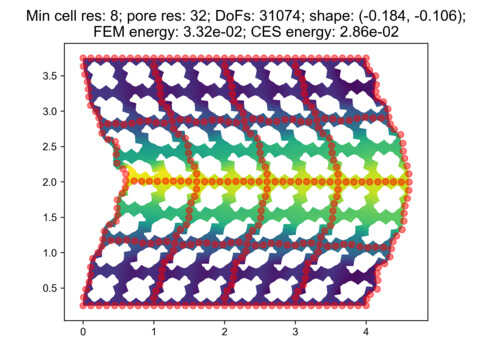}
\end{subfigure}
\newline
\begin{subfigure}{.45\textwidth}
  \centering
  \includegraphics[width=.8\linewidth]{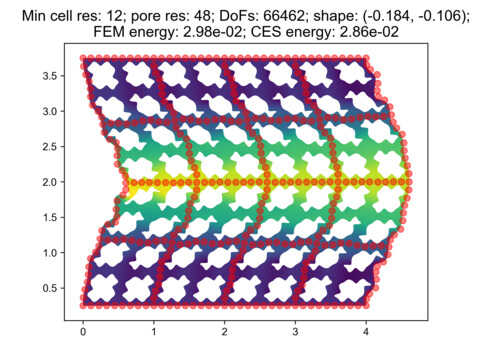}
\end{subfigure}
\begin{subfigure}{.45\textwidth}
  \centering
  \includegraphics[width=.8\linewidth]{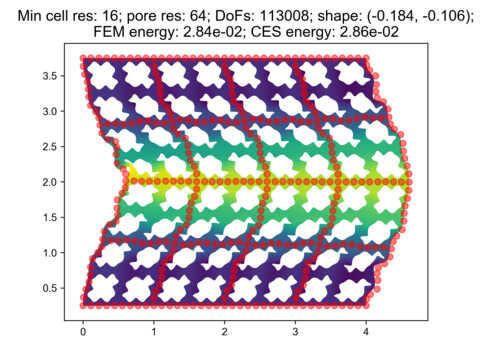}
\end{subfigure}
\end{figure}

\begin{figure}[H]
\begin{subfigure}{.45\textwidth}
  \centering
  \includegraphics[width=.8\linewidth]{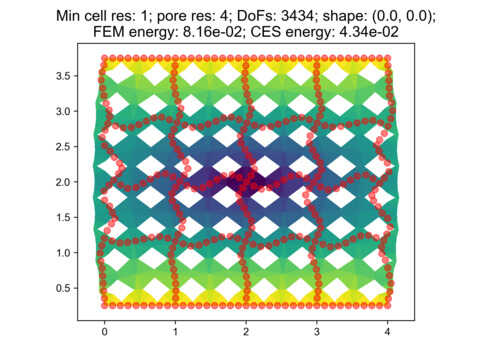}
\end{subfigure}
\begin{subfigure}{.45\textwidth}
  \centering
  \includegraphics[width=.8\linewidth]{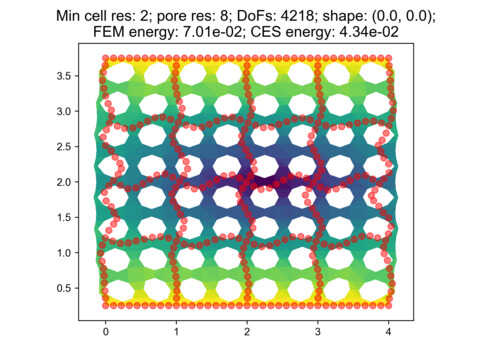}
\end{subfigure}
\newline
\begin{subfigure}{.45\textwidth}
  \centering
  \includegraphics[width=.8\linewidth]{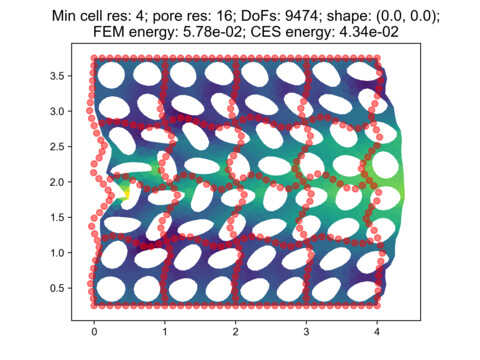}
\end{subfigure}
\begin{subfigure}{.45\textwidth}
  \centering
  \includegraphics[width=.8\linewidth]{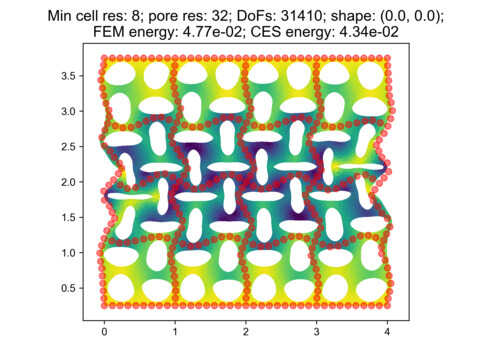}
\end{subfigure}
\newline
\begin{subfigure}{.45\textwidth}
  \centering
  \includegraphics[width=.8\linewidth]{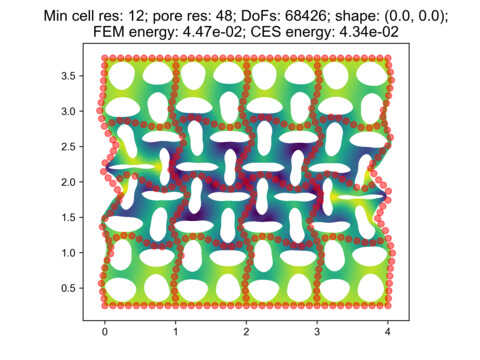}
\end{subfigure}
\begin{subfigure}{.45\textwidth}
  \centering
  \includegraphics[width=.8\linewidth]{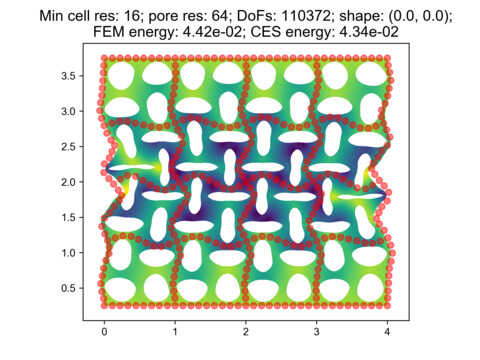}
\end{subfigure}
\end{figure}

\clearpage

\subsection{Tension}

\begin{figure}[H]
\begin{subfigure}{.45\textwidth}
  \centering
  \includegraphics[width=.8\linewidth]{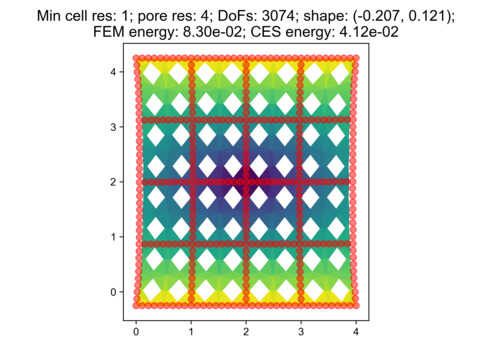}
\end{subfigure}
\begin{subfigure}{.45\textwidth}
  \centering
  \includegraphics[width=.8\linewidth]{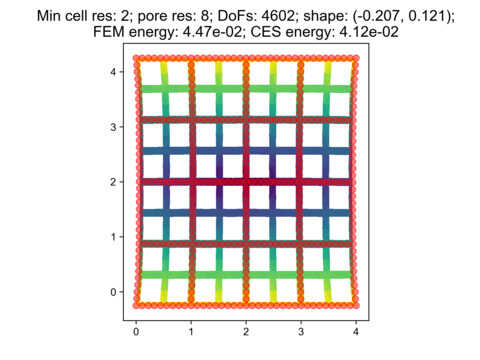}
\end{subfigure}
\newline
\begin{subfigure}{.45\textwidth}
  \centering
  \includegraphics[width=.8\linewidth]{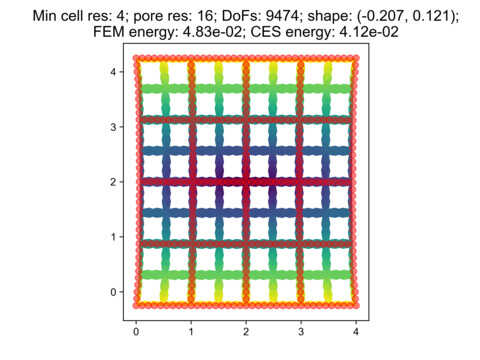}
\end{subfigure}
\begin{subfigure}{.45\textwidth}
  \centering
  \includegraphics[width=.8\linewidth]{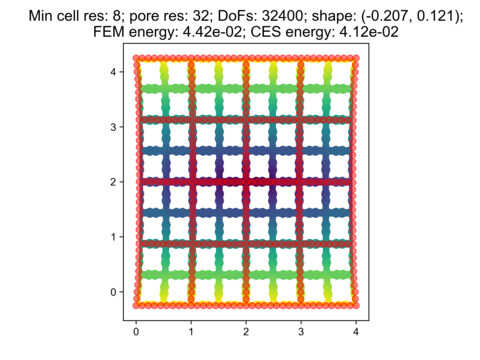}
\end{subfigure}
\newline
\begin{subfigure}{.45\textwidth}
  \centering
  \includegraphics[width=.8\linewidth]{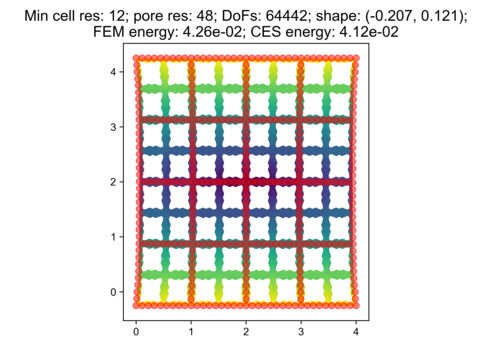}
\end{subfigure}
\begin{subfigure}{.45\textwidth}
  \centering
  \includegraphics[width=.8\linewidth]{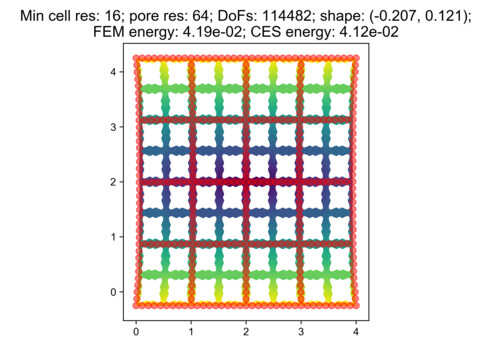}
\end{subfigure}
\end{figure}

\begin{figure}[H]
\begin{subfigure}{.45\textwidth}
  \centering
  \includegraphics[width=.8\linewidth]{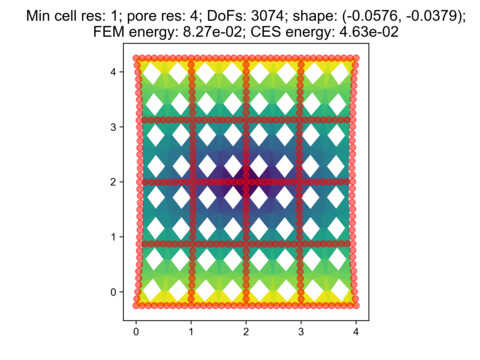}
\end{subfigure}
\begin{subfigure}{.45\textwidth}
  \centering
  \includegraphics[width=.8\linewidth]{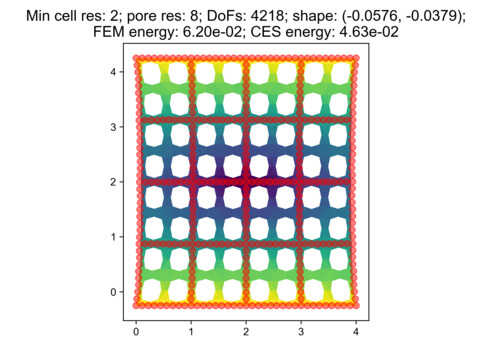}
\end{subfigure}
\newline
\begin{subfigure}{.45\textwidth}
  \centering
  \includegraphics[width=.8\linewidth]{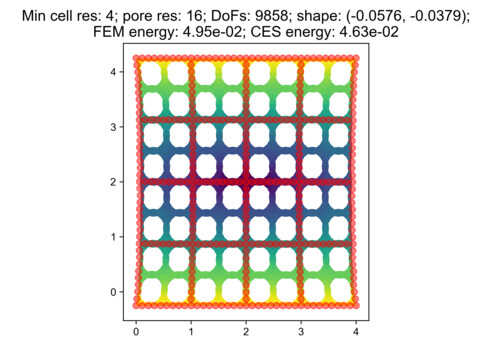}
\end{subfigure}
\begin{subfigure}{.45\textwidth}
  \centering
  \includegraphics[width=.8\linewidth]{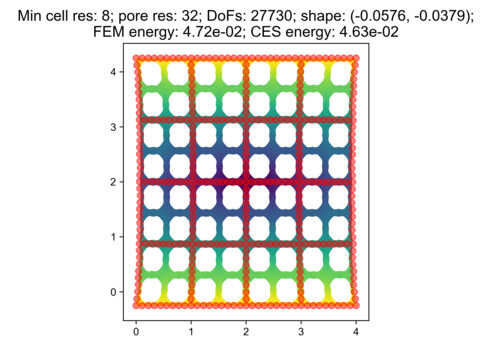}
\end{subfigure}
\newline
\begin{subfigure}{.45\textwidth}
  \centering
  \includegraphics[width=.8\linewidth]{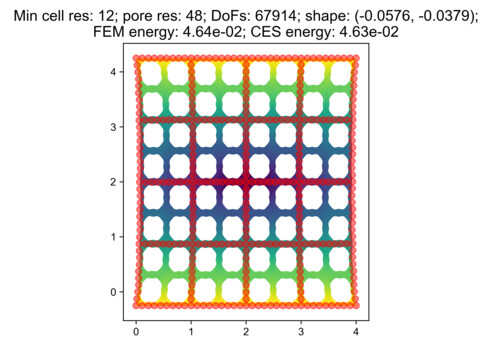}
\end{subfigure}
\begin{subfigure}{.45\textwidth}
  \centering
  \includegraphics[width=.8\linewidth]{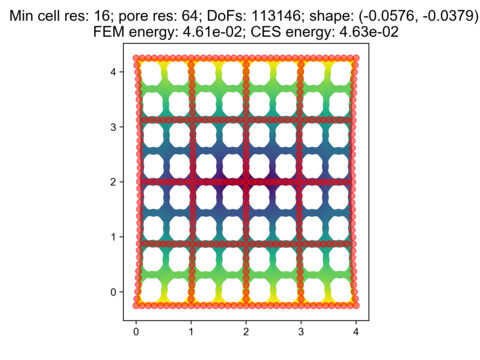}
\end{subfigure}
\end{figure}

\begin{figure}[H]
\begin{subfigure}{.45\textwidth}
  \centering
  \includegraphics[width=.8\linewidth]{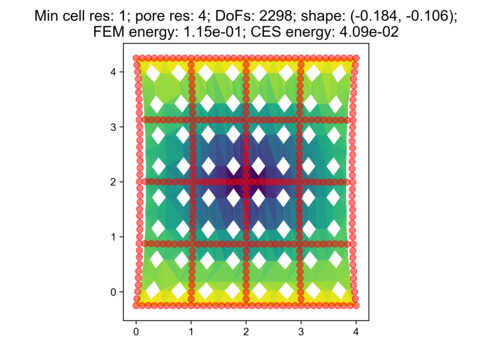}
\end{subfigure}
\begin{subfigure}{.45\textwidth}
  \centering
  \includegraphics[width=.8\linewidth]{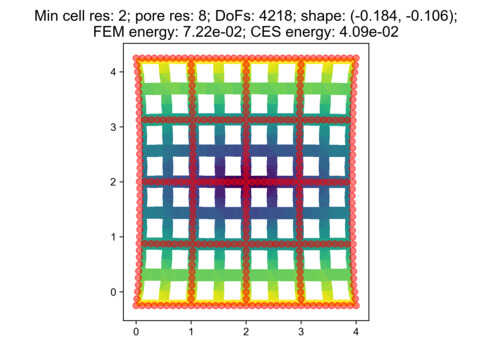}
\end{subfigure}
\newline
\begin{subfigure}{.45\textwidth}
  \centering
  \includegraphics[width=.8\linewidth]{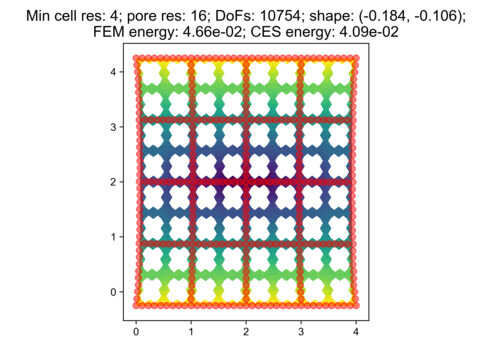}
\end{subfigure}
\begin{subfigure}{.45\textwidth}
  \centering
  \includegraphics[width=.8\linewidth]{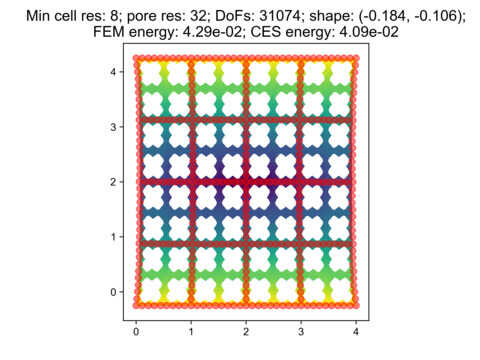}
\end{subfigure}
\newline
\begin{subfigure}{.45\textwidth}
  \centering
  \includegraphics[width=.8\linewidth]{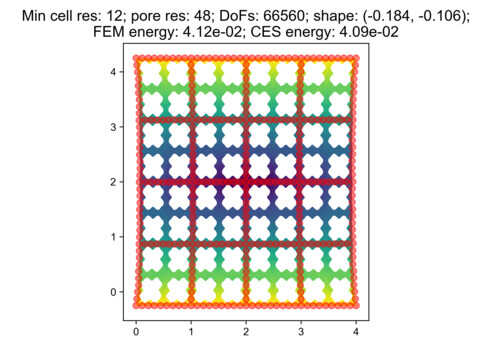}
\end{subfigure}
\begin{subfigure}{.45\textwidth}
  \centering
  \includegraphics[width=.8\linewidth]{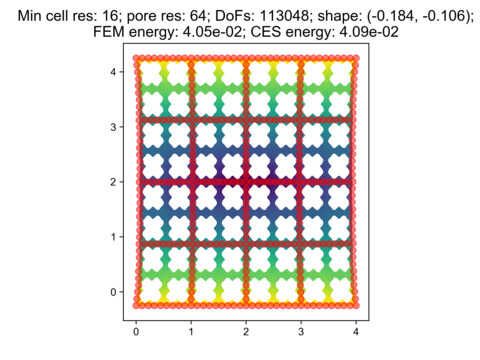}
\end{subfigure}
\end{figure}

\begin{figure}[H]
\begin{subfigure}{.45\textwidth}
  \centering
  \includegraphics[width=.8\linewidth]{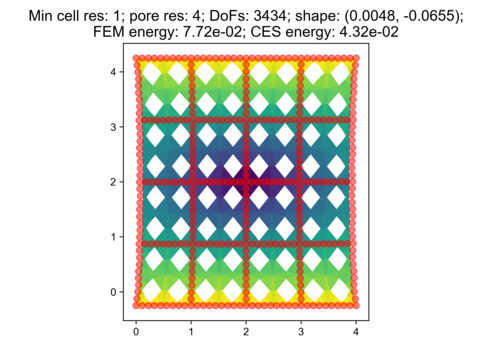}
\end{subfigure}
\begin{subfigure}{.45\textwidth}
  \centering
  \includegraphics[width=.8\linewidth]{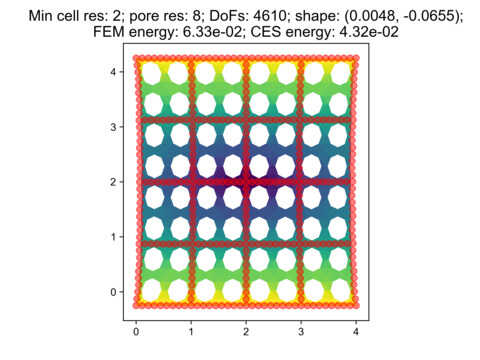}
\end{subfigure}
\newline
\begin{subfigure}{.45\textwidth}
  \centering
  \includegraphics[width=.8\linewidth]{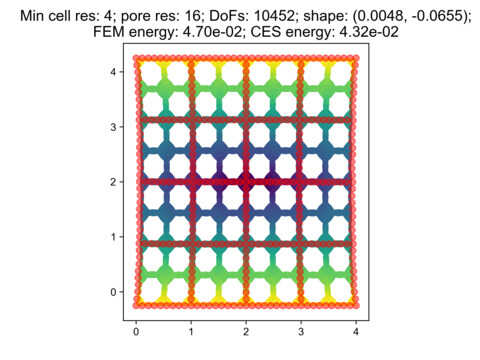}
\end{subfigure}
\begin{subfigure}{.45\textwidth}
  \centering
  \includegraphics[width=.8\linewidth]{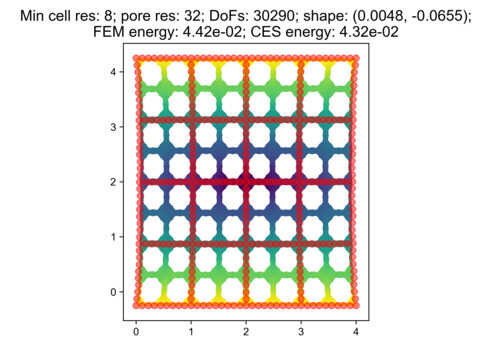}
\end{subfigure}
\newline
\begin{subfigure}{.45\textwidth}
  \centering
  \includegraphics[width=.8\linewidth]{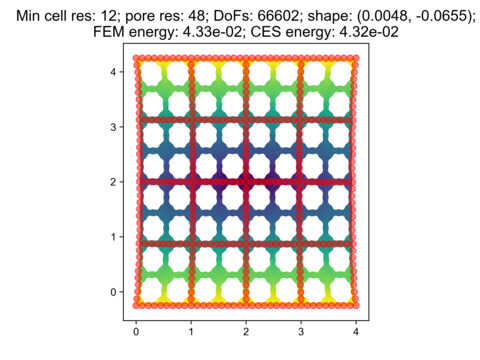}
\end{subfigure}
\begin{subfigure}{.45\textwidth}
  \centering
  \includegraphics[width=.8\linewidth]{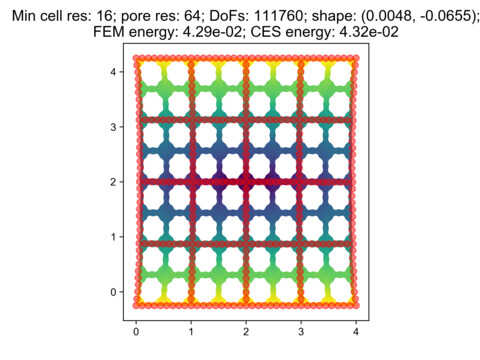}
\end{subfigure}
\end{figure}

\begin{figure}[H]
\begin{subfigure}{.45\textwidth}
  \centering
  \includegraphics[width=.8\linewidth]{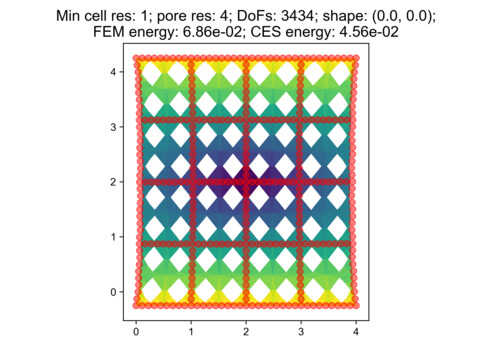}
\end{subfigure}
\begin{subfigure}{.45\textwidth}
  \centering
  \includegraphics[width=.8\linewidth]{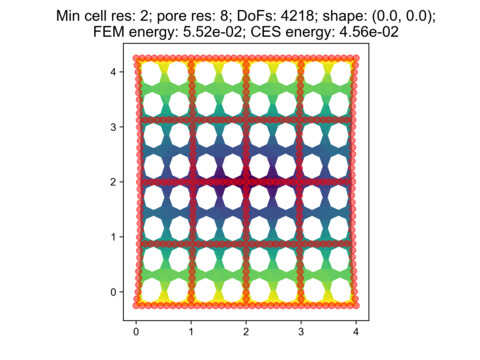}
\end{subfigure}
\newline
\begin{subfigure}{.45\textwidth}
  \centering
  \includegraphics[width=.8\linewidth]{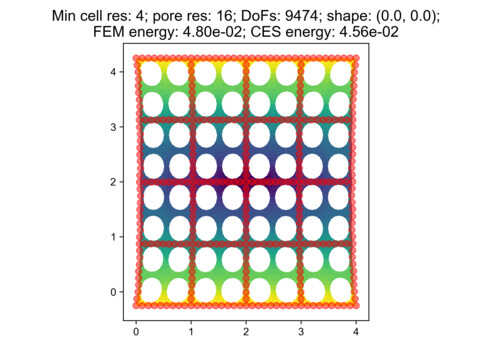}
\end{subfigure}
\begin{subfigure}{.45\textwidth}
  \centering
  \includegraphics[width=.8\linewidth]{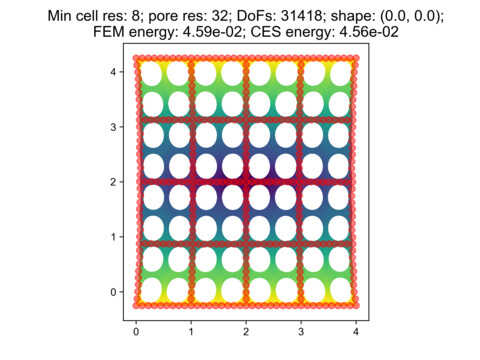}
\end{subfigure}
\newline
\begin{subfigure}{.45\textwidth}
  \centering
  \includegraphics[width=.8\linewidth]{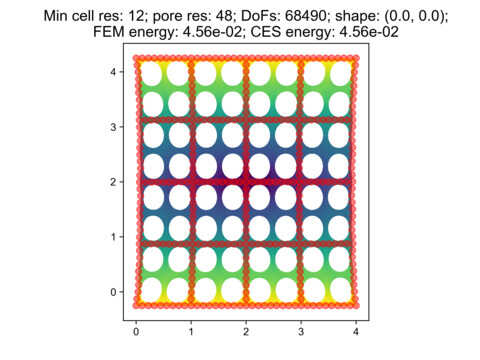}
\end{subfigure}
\begin{subfigure}{.45\textwidth}
  \centering
  \includegraphics[width=.8\linewidth]{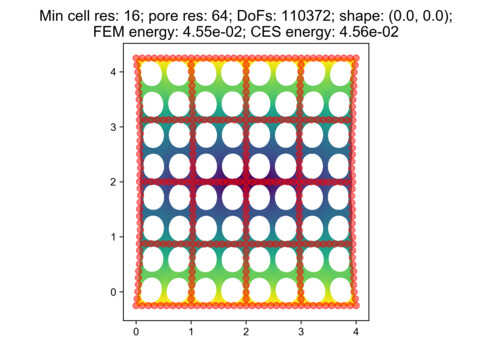}
\end{subfigure}
\end{figure}

\begin{figure}[H]
\begin{subfigure}{.45\textwidth}
  \centering
  \includegraphics[width=.8\linewidth]{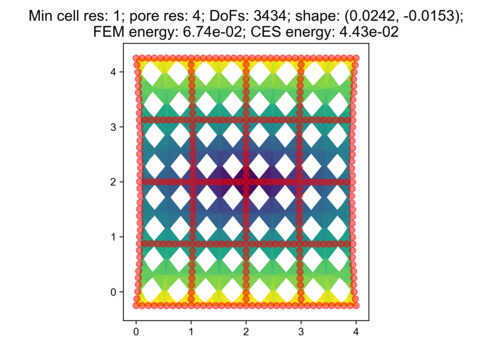}
\end{subfigure}
\begin{subfigure}{.45\textwidth}
  \centering
  \includegraphics[width=.8\linewidth]{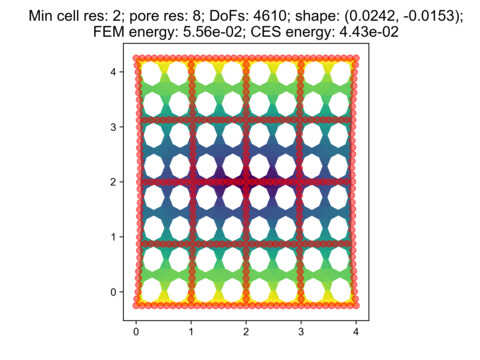}
\end{subfigure}
\newline
\begin{subfigure}{.45\textwidth}
  \centering
  \includegraphics[width=.8\linewidth]{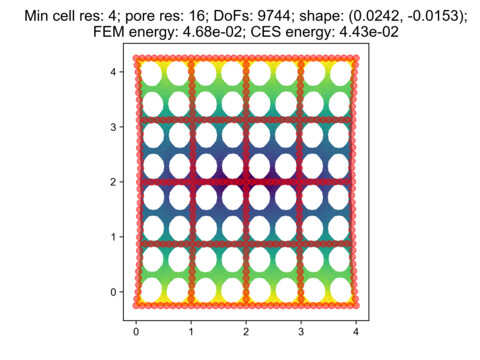}
\end{subfigure}
\begin{subfigure}{.45\textwidth}
  \centering
  \includegraphics[width=.8\linewidth]{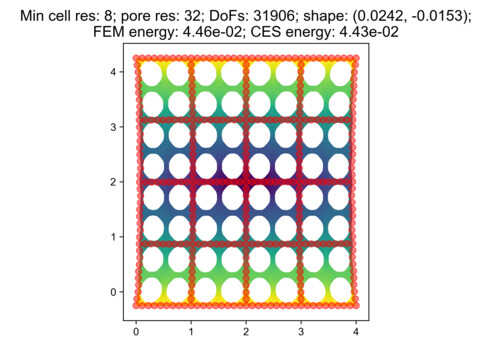}
\end{subfigure}
\newline
\begin{subfigure}{.45\textwidth}
  \centering
  \includegraphics[width=.8\linewidth]{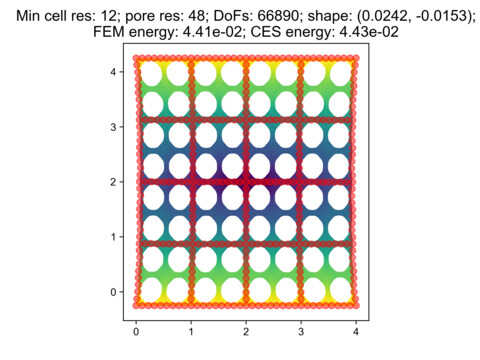}
\end{subfigure}
\begin{subfigure}{.45\textwidth}
  \centering
  \includegraphics[width=.8\linewidth]{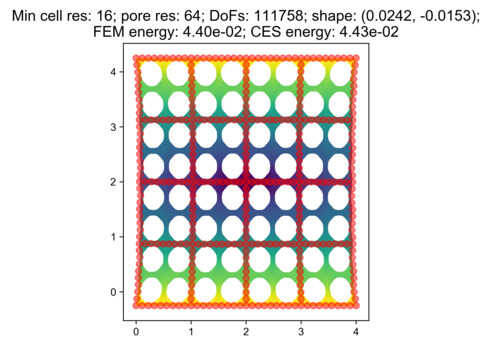}
\end{subfigure}
\end{figure}

\begin{figure}[H]
\begin{subfigure}{.45\textwidth}
  \centering
  \includegraphics[width=.8\linewidth]{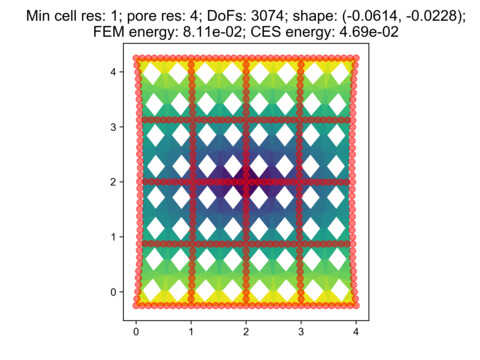}
\end{subfigure}
\begin{subfigure}{.45\textwidth}
  \centering
  \includegraphics[width=.8\linewidth]{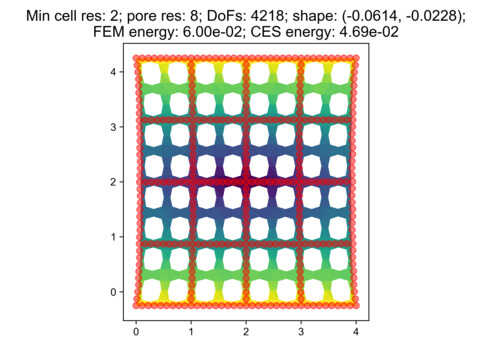}
\end{subfigure}
\newline
\begin{subfigure}{.45\textwidth}
  \centering
  \includegraphics[width=.8\linewidth]{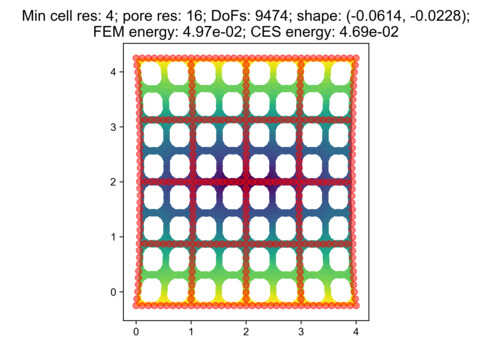}
\end{subfigure}
\begin{subfigure}{.45\textwidth}
  \centering
  \includegraphics[width=.8\linewidth]{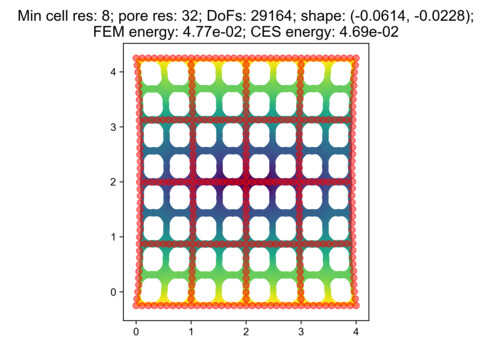}
\end{subfigure}
\newline
\begin{subfigure}{.45\textwidth}
  \centering
  \includegraphics[width=.8\linewidth]{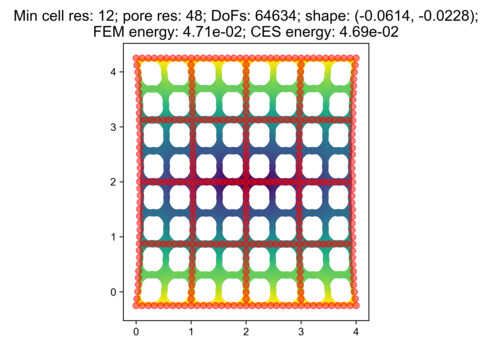}
\end{subfigure}
\begin{subfigure}{.45\textwidth}
  \centering
  \includegraphics[width=.8\linewidth]{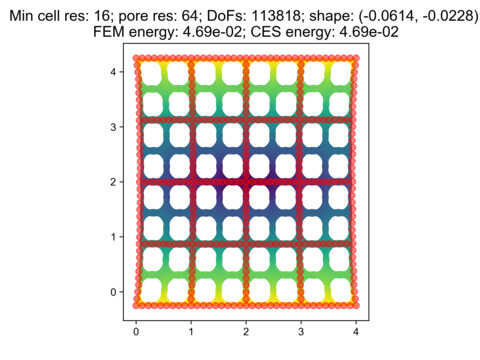}
\end{subfigure}
\end{figure}
